\documentclass[letterpaper]{article} 
\usepackage{aaai2026}  
\usepackage{times}  
\usepackage{helvet}  
\usepackage{courier}  
\usepackage[hyphens]{url}  
\usepackage{graphicx} 
\usepackage{natbib}  
\usepackage{caption} 
\usepackage{algorithm}
\usepackage{algorithmic}
\usepackage{amsmath}
\usepackage{booktabs}
\usepackage{subcaption}
\usepackage[table,xcdraw]{xcolor}
\usepackage{makecell}
\usepackage{amssymb}

\definecolor{textcolorone}{RGB}{139, 64, 0}
\definecolor{textcolortwo}{RGB}{0,0,128}
\newcommand{\colorone}[1]{\textcolor{textcolorone}{#1}} 
\newcommand{\colortwo}[1]{\textcolor{textcolortwo}{#1}} 

\usepackage{newfloat}
\usepackage{listings}
\DeclareCaptionStyle{ruled}{labelfont=normalfont,labelsep=colon,strut=off} 
\floatstyle{ruled}
\newfloat{listing}{tb}{lst}{}
\floatname{listing}{Listing}
\title{BATIS: Bayesian Approaches for \\ Targeted Improvement of Species Distribution Models}
\author{
    Catherine Villeneuve\textsuperscript{\rm 1, 2}, 
    Benjamin Akera\textsuperscript{\rm 1, 2},
    Mélisande Teng\textsuperscript{\rm 1,3},
    David Rolnick\textsuperscript{\rm 1,2}
}
\affiliations{
    \textsuperscript{\rm 1}McGill University,
    \textsuperscript{\rm 2}Mila–Quebec AI Institute,
    \textsuperscript{\rm 3}Université de Montréal \\ 
    catherine.villeneuve@mila.quebec
}

\begin{document}

\maketitle

\begin{abstract}
Species distribution models (SDMs), which aim to predict species occurrence based on environmental variables, are widely used to monitor and respond to biodiversity change. Recent deep learning advances for SDMs have been shown to perform well on complex and heterogeneous datasets, but their effectiveness remains limited by spatial biases in the data. In this paper, we revisit deep SDMs from a Bayesian perspective and introduce BATIS, a novel and practical framework wherein prior predictions are updated iteratively using limited observational data. Models must appropriately capture both aleatoric and epistemic uncertainty to effectively combine fine-grained local insights with broader ecological patterns. We benchmark an extensive set of uncertainty quantification approaches on a novel dataset including citizen science observations from the eBird platform. Our empirical study shows how Bayesian deep learning approaches can greatly improve the reliability of SDMs in data-scarce locations, which can contribute to ecological understanding and conservation efforts. 
\end{abstract}

 \begin{links}
\link{Code}{https://github.com/cath34/batis}
\link{Datasets}{https://huggingface.co/datasets/cathv/BATIS}
\link{Extended version}{https://arxiv.org/abs/2510.19749}
\end{links}

\section{Introduction}
Understanding the factors governing the spatial distribution of species is a long-standing topic of interest in ecology \cite{decandolle, pollock2015}. The rapid loss of biodiversity further highlights the need to better understand species distributions and conservation status, which requires robust modeling methods \cite{nospeciesleftbehind}. \textit{Species distribution models} (SDMs) are at the forefront of this research, enabling ecologists to predict and understand species occurrence  by linking  observational data (presence or absence) to environmental covariates \cite{elithsdm}. These models have proved to be essential tools to support conservation planning and management \cite{sdmessentialtool}, with notable examples including predicting species range shift \cite{rangeshiftpaper}, mapping invasive species risk \cite{alienspecies}, supporting translocation programs \cite{translocpaper}, reducing bycatch rates in fisheries \cite{fisheriespaper}, and informing IUCN Red List assessments \cite{iucnpaper}.

SDMs are traditionally based on statistical approaches, offering strong interpretability and theoretical grounding. However these methods often rely on restrictive assumptions such as linearity, leading them to struggle with high-dimensional datasets. Recently, deep learning approaches to SDMs proposed to overcome these challenges and were shown to consistently outperform statistical ecology methods for large and heterogeneous datasets \cite{Deneu_Servajean_Bonnet_Botella_Munoz_Joly_2021}. 

\begin{figure}[h!]
    \centering    \includegraphics[width=\linewidth]{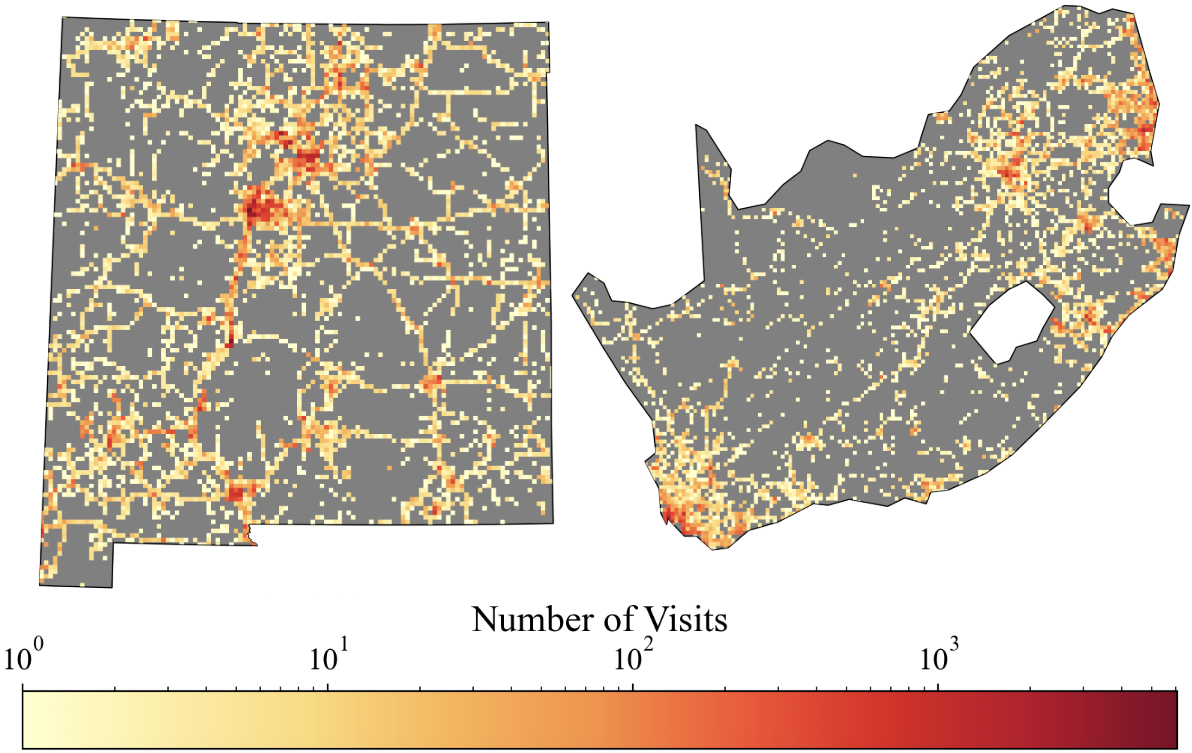}
    \caption{Distribution of  citizen science sampling trips across New Mexico, US (left) and South Africa (right), retrieved from the eBird database for the whole year of 2024. The value assigned to each 5km\textsuperscript{2} (US) and 10km\textsuperscript{2} (South Africa) grid cell corresponds to the total number of visits that were registered within that cell. } \label{figintro}
\end{figure}

Although a wide variety of SDM approaches have now been introduced, their effectiveness at large spatio-temporal scales remains hindered by the abundance and geographical coverage of the data used to train them \cite{minimumsdmdata}. In most cases, observations are disproportionately concentrated in areas easily accessible by roads \cite{roadbias} and with high human population densities \cite{densitybias}. In contrast, remote locations are most often associated with very limited to no observations, as illustrated in Fig.~\ref{figintro}.

Data distribution biases especially affect machine learning-based approaches, which typically rely on trustworthy labeled data. Limited observations in a given location can lead to poor estimates of species occurrence \cite{finkpaperencounterate} -- essentially, noisy labels -- and na\"ively incorporating such data points in model training can degrade performance. However, sparse observations still offer potentially valuable information, and finding ways to properly leverage it could help improve SDMs and guide conservation efforts in remote or underresourced areas. We argue that uncertainty-aware methods can offer a principled way of leveraging limited location-specific observations, enabling controlled prediction updates that can balance global-scale models with local-scale observations.

In this work, we tackle the spatial heterogeneity challenge of biodiversity data by revisiting species distribution modeling from a Bayesian perspective, introducing \textbf{BATIS}, a framework enabling the application of \textbf{B}ayesian \textbf{A}pproaches for \textbf{T}argeted \textbf{I}mprovement of \textbf{S}pecies distribution models (named after the batis, a bird found in sub-Saharan Africa). In this framework, prior predictions from ML are iteratively refined offline with limited field observations in data-scarce locations, testing the ability of ML models to capture both the epistemic (model-inherent) and aleatoric (data-inherent) uncertainty of SDM data. Quantifying epistemic uncertainty provides a measure of the model’s confidence in its prior predictions and highlights how informative additional field observations could be, while measuring aleatoric uncertainty ensures that the noise associated with the additional data is appropriately taken into account. We present an empirical study of a variety of uncertainty-aware ML approaches approaches on this task, using a novel large-scale dataset. 
\begin{itemize}
    \item We introduce the framework of iteratively refining the prior predictions of an uncertainty-aware SDM with the help of limited additional on-the-ground information. 
    \item We benchmark a variety of state-of-the-art uncertainty estimation ML methods on the above framework, and present a novel dataset derived from eBird  \cite{ebirdpaper}, building on the SatBird \cite{NEURIPS2023_ef7653bb} dataset.
    \item We find that our framework can rapidly improve predicted species distribution models in data-scarce locations, even with minimal additional ground truth data ($<$10 samples), effectively combining broader ecological patterns with fine-grained, location-specific insights. 
    \item We provide evidence that machine learning approaches relying on aleatoric uncertainty are the most effective at improving predictions in  low-data regimes.
\end{itemize}
BATIS can be seen as an innovative way of coupling the predictive power of modern deep learning to the statistical rigor of traditional hierarchical inference, thereby creating a bridge between ML and established methods in ecology. Our code and  datasets are available open-source.
\section{Related Works}
\paragraph{Machine Learning for SDMs.} Species distributions have been widely estimated from environmental data with statistical ecology methods. Popular SDMs in ecology include techniques based on maximum entropy (MaxEnt) \cite{maxenticml}, random forests \cite{rfsdm}, and generalized linear models \cite{GUISAN200289}.
More recently, a wide variety of deep learning approaches have been introduced, including simple multi-layer perceptrons \cite{zbindenpresenceabsence}, computer vision methods leveraging remote sensing imagery \cite{Estopinan_Servajean_Bonnet_Munoz_Joly_2022, NEURIPS2023_ef7653bb} or location embeddings \cite{SINR_icml23}, and even language-based reasoning methods \cite{NEURIPS2024_1f96b24d}. Active learning \cite{NEURIPS2023_82eec786} and few-shot learning \cite{fewshotpaper} frameworks have also been introduced. Deep learning models have been shown to consistently outperform traditional approaches in ecology \cite{Deneu_Servajean_Bonnet_Botella_Munoz_Joly_2021}, notably when applied to large-scale citizen science datasets \cite{gomescitizen}. 

Some deep learning-based SDMs have focused on estimating binary species range maps \cite{dorm2024generatingbinaryspeciesrange}. This corresponds to a multi-label classification problem in which the list of species that \textit{can} be encountered at a given location is predicted, regardless of the frequency at which they are observed. Another important challenge is \emph{encounter rate estimation}, i.e.~a multi-output regression problem aimed at predicting the
average rate at which observers encounter species at a given location \cite{finkebird1}. The use of ML for predicting encounter rates has been explored in \citet{NEURIPS2023_ef7653bb} and \citet{abdelwahed2024predictingspeciesoccurrencepatterns}, a line of work which we follow here. Encounter rates are closely linked to the notion of \textit{occupancy}, i.e.~the probability of a species occupying a given site. Occupancy is a core concept in ecology \cite{MacKenzie_2018}, and understanding occupancy patterns is essential for prioritizing conservation efforts. Because encounter rates can be seen as the product of occupancy and detection probability \cite{finkebird1}, they can provide a more nuanced and realistic representation of species distributions than binary range maps. 
\paragraph{Bayesian Inference for SDMs.} In statistical ecology, SDMs have been estimated in a Bayesian context using computationally intensive algorithms, such as Markov Chain Monte Carlo (MCMC) approaches \cite{Zulian_Miller_Ferraz_2021} and integrated nested Laplace approximation \cite{Omre_Fjeldstad_Forberg_2024}. \citet{Golding_Purse_2016} proposed a more flexible approach based on Gaussian processes, but this was applicable only to a limited number of species and covariates. Our work is the first to reframe deep learning-based SDMs under a Bayesian perspective. 
\paragraph{Uncertainty Quantification for SDMs.} Uncertainty quantification is recognized by ecologists as a major challenge for SDMs \cite{importantuncertaintysdm}. To date, however, the vast majority of works do not investigate the effects of uncertainty on model outputs \cite{paperodmap}. Only a limited number of statistical tools have been proposed to measure and incorporate uncertainty estimates into SDMs \cite{uncsdm1}, and these tools are not easily scalable, as they often require computationally expensive MCMC iterations to estimate posterior probability distributions \cite{Rocchini_Marcantonio_Arhonditsis_Cacciato_Hauffe_He_2019}. Thus, they are only appropriate for SDMs relying on a very small number of covariates and species. Approximation techniques such as model ensembling \cite{GUO201567} have been applied to quantify uncertainty in SDMs \cite{CONVERTINO20121}, but their use remain limited. Recently, an active learning framework was proposed to sequentially select geographic locations that would best reduce uncertainty of an SDM for previously unmapped species \citet{NEURIPS2023_82eec786}. This work however does not take a Bayesian approach to improving predictions, instead estimating uncertainty via disagreement across an ensemble of models to guide the active learning step.
\paragraph{Uncertainty Quantification in Deep Learning.} An extensive variety of approaches have been proposed to approximate uncertainty in the Bayesian deep learning literature, since computing a posterior distribution over the parameters of a
neural network according to the rule of Bayesian inference is analytically intractable \cite{benchmarkdiabetes}. However, these approaches have mostly been tested on generic vision benchmarks \cite{NEURIPS2024_5afa9cb1} such as CIFAR-10 \cite{Krizhevsky2009LearningML} and ImageNet-1k \cite{5206848}, which do not reflect the complexity of real-world tasks. Our work is the first to integrate such approaches for solving a complex ecological task. 
\section{The BATIS Framework}
\label{noveltasksection}
\subsection{Species Encounter Rate Estimation}
We here formalize the problem of species distribution modeling in the context of checklist-based surveys (presence-absence data) from ecologists and/or citizen science platforms. Let $L$ be a region of interest, and $\mathcal{S} = \{ s_{1}, \dots, s_{N} \}$ be a set of $N$ species that may potentially exist within $L$. We consider observations within $L$ as occurring at one of a set of $K$ discrete \emph{hotspots} $\mathcal{H} = \{ h_{1}, \dots, h_{K} \}$, each representing an area in which observers can record species (e.g.~a park, geographic feature, or address). To each hotspot $h_{k} \in \{ 1, \dots, K \}$ is associated a set of checklists $\mathcal{C}_{k} = \{ \mathbf{c}_{j} \}_{j=1}^{|\mathcal{C}_{k}|}$, where each checklist 
$\mathbf{c}_{j} \in \{ 0, 1 \}^{N}$ indicates which of the $N$ species are recorded as present (1) or absent (0) during an observer's visit to the location. For any (non-empty) set $\mathcal{C}_{k}$ we define $p_{i, k}$ of species $i$ at hotspot $k$ to be the rate at which $i$ is reported across checklists:
\begin{equation}
    p_{i, k} := \frac{1}{|\mathcal{C}_{k}|} \sum_{\mathbf{c}_{j} \in \mathcal{C}_{k}} \mathbf{c}_{i, j}
\end{equation}
It is a standard assumption that as $|\mathcal{C}_k|\to\infty$ (assuming a diversity of observers, observation dates, etc.), the value $p_{i,k}$ converges to an ecological quantity, the \emph{encounter rate} $p^\infty_{i,k}$, representing the chance that a random observer will encounter species $i$ at $h_k$ on any given occasion.

Following \citet{NEURIPS2023_ef7653bb,abdelwahed2024predictingspeciesoccurrencepatterns}, we consider a machine learning formulation of species distribution modeling in which the encounter rate $p^\infty_{i,k}$ is to be estimated for each species $i$, given a set of information associated with the hotspot $h_k$ -- e.g., average precipitation, temperature, etc., as well as satellite imagery of the location. In practice, of course, the ground truth value $p^\infty_{i,k}$ is not directly observable, but is well-approximated by $p_{i,k}$ for hotspots $h_k$ where $|\mathcal{C}_k|$ is sufficiently large; therefore the task is to estimate $p_{i,k}$.
\subsection{Bayesian Estimates of Encounter Rates}
For hotspot $h_k \in \mathcal{H}$, let $C_{i, k} = p_{i,k}\cdot |\mathcal{C}_{k}|$ be the number of times species $i$ was observed at $h_k$. Then, we may assume that $C_{i, k}$ follows a binomial distribution parameterized by the true encounter rate -- i.e.~$C_{i, k} \sim \text{Binomial}(J, p^{\infty}_{i, k})$. This simple assumption allows us to formulate a \emph{Bayesian inference problem}, in which we place a \emph{Beta prior} over $p^{\infty}_{i, k}$: 
\begin{equation}
    p^{\infty}_{i, k} \sim \text{Beta} (\alpha_{i,k}, \beta_{i,k})
\end{equation}
Instead of estimating $p^{\infty}_{i, k}$ directly, we may consider instead the problem of estimating the parameters $\alpha_{i,k}$ and $\beta_{i,k}$, which can be defined by the closed-form equations
\begin{equation}
      \alpha_{i, k} = \mu_{i,k} \left ( \frac{\mu_{i,k} (1-\mu_{i,k} )}{\sigma^{2}_{i,k}} - 1 \right )
\end{equation}
\begin{equation}
    \beta_{i, k} = (1 -\mu_{i,k} )\left ( \frac{\mu_{i,k} (1-\mu_{i,k} )}{\sigma^{2}_{i,k}} - 1 \right )
\end{equation}
where at each point $\mu_{i, k} \in [0, 1]$ encodes our best estimate of the encounter rate of species $i$ at $h_k$, and $\sigma^{2}_{i,k} \leq \mu_{i, k} (1 - \mu_{i, k})$ encodes our uncertainty about that belief.

While previous approaches of predicting species distributions using environmental variables made it possible to estimate encounter rates at a location without incorporating observations at the location, our Bayesian framework makes it possible to combine broad patterns learned by the SDM with any individual checklists recorded at the location in question. Namely, we may consider the output of an SDM to be an initial estimate of $\alpha_{i,k}$ and $\beta_{i,k}$, and we may then incorporate additional information from a checklist of observations by performing a Bayesian update to these parameters to obtain the posterior. In the Beta distribution, this update takes a convenient form: $\alpha_{i, k} :=  \alpha_{i, k} + \sum_{\mathbf{c}^{\prime}_{j} \in \mathcal{C}^{\prime}_{k}} \mathbf{c}^{\prime}_{i, j}$ and $\beta_{i, k} := \beta_{i, k} + 1 - \sum_{\mathbf{c}^{\prime}_{j} \in \mathcal{C}^{\prime}_{k}} \mathbf{c}^{\prime}_{i, j}$. 

Our Bayesian reformulation of the species distribution modeling problem thus allows us to iteratively update prior encounter rates predictions coming from an ML model with additional field observations as they are recorded. 
\subsection{Applications}
Our approach can significantly improve encounter rate predictions in data-scarce regions (see section \ref{section:results}). Specifically, our framework is useful since, due to the geographic distribution of species observational data, most locations have a small but meaningful number of observations -- that is, there are insufficient observations to reliably calculate the encounter rate empirically, yet enough observations that they contain valuable information. BATIS is particularly well-suited for location-specific ecological questions, such as assessing changes in habitat use and monitoring threatened species presence in protected areas. Moreover, because it allows for fast and lightweight updates of prior predictions, our framework is ideal for integrating daily or weekly updates from citizen science databases such as eBird or iNaturalist \cite{iNaturalist}, which are continuously growing, into the SDMs used to inform conservation and land use decisions.
\section{The BATIS Benchmark}
\label{sec:benchmark}
\subsection{Dataset}
We introduce a dataset for the BATIS framework. We rely on the eBird citizen science database \cite{ebirdpaper}, containing  millions of bird observations across the globe standardized in the form of \textit{checklists}, indicating which species were seen or not during a given survey trip. This large amount of data allows us to create a benchmark of test cases that mirrors the real-world scenarios in which BATIS can be applied. We can notably simulate data-deficient locations by providing partial information to SDM algorithms so as to test their generalization capabilities against full ground-truth data. 
 Although eBird strictly focuses on birds, its structure and scale makes it an ideal setting for evaluating BATIS methods,  which we hope may then be applied to less standardized datasets, such as those available through GBIF.

Following \citet{NEURIPS2023_ef7653bb}, we first extracted complete checklists from all hotspots associated with the mainland portions of Kenya (KE), South Africa (ZA), and the contiguous United States. For the US, we followed the same approach as \citet{NEURIPS2023_ef7653bb}, splitting the dataset into two seasons: summer (US-S, breeding period) and winter (US-W, non-breeding period). We did not consider seasonality for the KE and ZA datasets, as these regions have significantly fewer available checklists and seasonal migrations are also less pronounced in KE and ZA than in the US. Table \ref{tab:composition} summarizes the composition of each subdataset. Further details on the dataset can be found in Appendix A.
\begin{table}[h]
\centering 
\setlength{\tabcolsep}{1mm}{ 
\fontsize{9pt}{\baselineskip}
\begin{tabular}{lcccc}
\hline \toprule
\multicolumn{1}{c}{}          & \textbf{KE} & \textbf{ZA} & \textbf{US-W} & \textbf{US-S} \\ \hline \toprule
\textbf{Start Date}           & 01/01/10    & 01/01/18    & 01/12/22      & 01/06/22      \\
\textbf{End Date}             & 31/12/23    & 17/06/24    & 31/01/23      & 31/07/22      \\
\textbf{N. Species}    &  1,054          &   755          &   670            &   670            \\
\textbf{N. Hotspots}   &  8,551           & 6,643            &    45,882           &   98,443            \\
\textbf{N. Checklists} &   44,852          &  498,867           &  3,673,742             &  3,920,846             \\ \hline \toprule
\end{tabular}
}
\caption{Summary of the composition of the four subdatasets}
\label{tab:composition}
\end{table}

\paragraph{Remote Sensing Variables.} Our dataset also includes bioclimatic rasters from the WorldClim2.1 model (1km resolution) and Sentinel-2 satellite images (10m resolution) for each hotspot. These serve as predictive variables (inputs) to the models we test, and were selected based on the work of \citet{NEURIPS2023_ef7653bb}.  More information on these variables can be found in Appendix A. 

\paragraph{Splits.} We reserved 50\% of the hotspots with $\geq$ 15 checklists for the test set (15 being a large enough value to reliably estimate the true encounter rate $p^{\infty}_{i,k}$). The remaining hotspots were split into training and validation sets in a 80:20 ratio, and we used DBSCAN to cluster hotspots and proportionally distribute them across splits, which prevents auto-correlation and over-fitting  \cite{autocorrgis}. Table 5 (Appendix A) summarizes the content of each split.
\subsection{Uncertainty-Agnostic Methods}
We considered four SDM baselines in our benchmark. The first one, referred to as \textbf{Mean Encounter Rate}, simply predicts the average encounter rate on the training set for each species, ignoring the inputs. We also include a \textbf{Random Forest} baseline trained from bioclimatic variables only, as this is a frequently used approach in ecology. Because bioclimatic variables are still almost always exclusively used in ecology, we include a \textbf{Multi-Layer Perceptron Model (MLP)} solely trained from these inputs. Finally, we consider the \textbf{ResNet-18} approach introduced by \citet{NEURIPS2023_ef7653bb}, as it was shown to be the best performing model among the SatBird benchmark. The ResNet-18 relies on both bioclimatic variables and satellite imagery as inputs. More information on each of our SDM baselines can be found in Appendix F. 
\subsection{Uncertainty-Aware Methods}
\label{uncertaintyquantifsection}
We considered the uncertainty estimation approaches listed below. Except for the Fixed and Historical Variance baselines, we restricted ourselves exclusively to distributional approaches, as they offer a more principled way of estimating both the mean and variance of each predicted encounter rate \cite{NEURIPS2024_5afa9cb1}. More details on each approach can be found in Appendix F.

\paragraph{Fixed Variance.} Let $\widehat{\mu}$ be the encounter rate predicted by an SDM.  Our Fixed Variance (FV) baseline directly considers $\widehat{\mu}$ as the mean, and fixes a pre-determined variance $\widehat{\sigma^{2}}$ to initialize the prior. The pre-determined variance must be contained within $\widehat{\sigma^{2}} \leq \widehat{\mu}(1-\widehat{\mu})$ to maintain a valid Beta distribution. FV also includes a hyperparameter $\tau \in (0, 1]$ to finetune $\widehat{\sigma^{2}}$ more easily:
\begin{equation}
\widehat{\sigma^{2}} = \tau \cdot \widehat{\mu}(1-\widehat{\mu}).
\end{equation}
We considered $\tau = 1$ in our experiments, as we were interested in investigating the behavior of FV when the non-informative $\text{Beta}(0, 0)$ distribution is assigned as the prior.
\paragraph{Historical Variance.} Our Historical Variance (HV) baseline computes $\widehat{\sigma^{2}}$ from past checklists instead, which can be viewed as a way of trying to integrate the uncertainty associated with the data collection process into our framework. In edge cases where $\widehat{\sigma^{2}} > \widehat{\mu}(1-\widehat{\mu})$, the variance is instead fixed at $\widehat{\sigma^{2}} = \widehat{\mu}(1-\widehat{\mu})$. In our experiments, we limited the number of past checklists used to compute the prior variance to 5, as we are interested in how this method performs in a low-data regime.
\paragraph{Monte-Carlo Dropout \cite{pmlr-v48-gal16}.} MC Dropout (MCD) can be shown to be a variational approximation of a deep Gaussian process \cite{pmlr-v48-gal16}. A proportion of neurons are randomly deactivated at training and test time, and $M$ forward passes are used to compute mean and variance for each location. MCD is primarily effective at capturing epistemic uncertainty.  

\begin{figure*}[h!]
    \centering
\includegraphics[width=\linewidth]{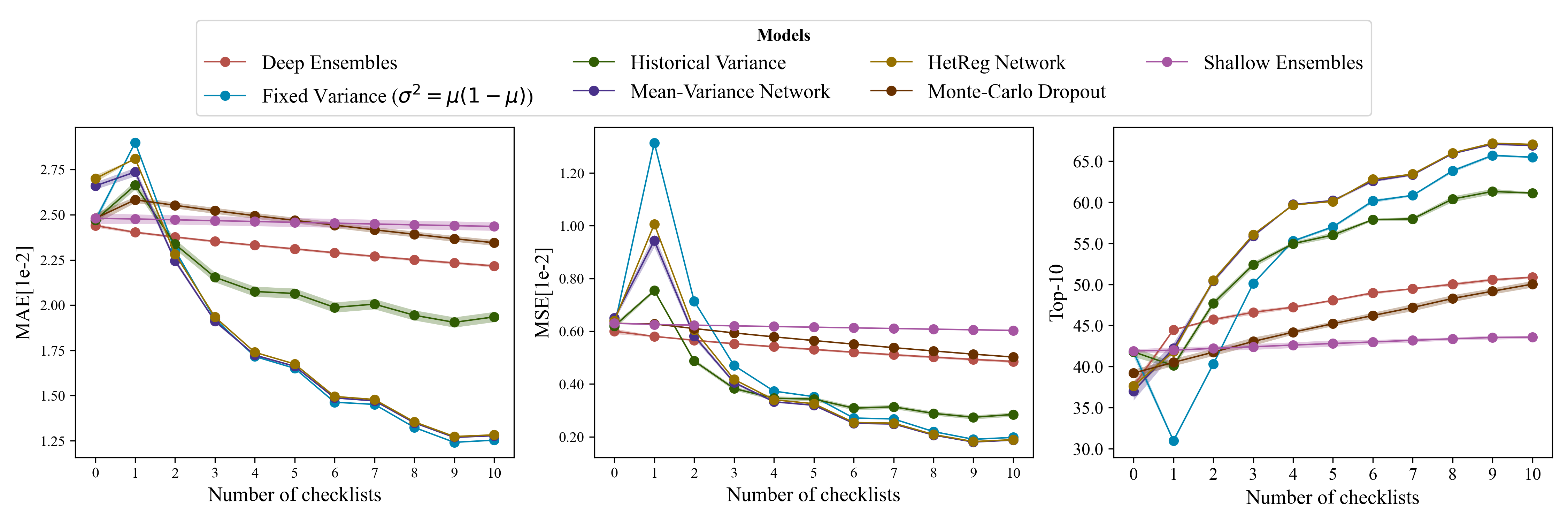}
 \caption{Iterative improvements for the different uncertainty estimation approaches with increasing number of checklist updates for the MAE, MSE and Top-10 metrics on the South Africa Region test set. We report the mean on three seeds and standard deviations for each model.}
 \label{figresiterative}
\end{figure*}

\paragraph{Deep Ensembles \cite{deepens}.} In Deep Ensembles (DE), an ensemble of $M$ neural networks are independently trained on the same task, and their individual predictions on the same input can be used to estimate mean and variance. Despite its simplicity, DE is the current state-of-the-art for estimating epistemic uncertainty in deep learning \cite{mukhoti2021deterministic}. 

\paragraph{Shallow-Ensembles \cite{lee2015mheadsbetterone}. } Shallow Ensembles (SE) aim to offer a computationally cheaper alternative to DE by using a shared backbone and $M$ output heads. The mean and variance of each predictions can be estimated by averaging the outputs of the $M$ heads.

\paragraph{Mean-Variance Network \cite{mvapaper2, mvapaper1} .} A Mean-Variance Network (MVN)  maps each location to two outputs: a predicted mean encounter rate vector and a predicted variance vector. The predicted variance can be considered as a measure of aleatoric uncertainty, as it aims to reflect the inherent task difficulty. MVNs are trained using the Gaussian negative log-likelihood loss function, presuming that encounter rates are normally distributed, which results in an increase in predicted variance when the predicted mean encounter rate differs greatly from the ground truth value.
\paragraph{Heteroscedastic Regression Neural Network \cite{hetregpaper}.} A Heteroscedastic Regression Neural Network (HetReg) is very similar to an MVN, but adds MC Dropout sampling to simultaneously quantify aleatoric and epistemic uncertainty. $M$ dropout passes are used to compute the mean encounter rate vector, and variance is estimated by adding epistemic uncertainty (variance computed from $M$ predicted encounter rates) to aleatoric uncertainty (mean computed from $M$ predicted variances). 
\subsection{Experiments}

We performed the experiments described below. Training protocol details, including hyperparameters, and compute resources can be found in Appendix G.  
\paragraph{Overall performance.} We trained each model on each of the four subdatasets. We report baseline performance for each model in Table \ref{bigmegatable}. Then, we investigate how the framework introduced in section \ref{noveltasksection} with the uncertainty quantification approaches described in section \ref{uncertaintyquantifsection} can improve these baseline results. We report performance after updating the posterior distribution of each hotspot with five new checklists for each model.   
\paragraph{Region-specific episodic behavior.} We investigate the episodic behavior of our proposed Bayesian framework for the MLP and ResNet-18 models. We study how average performance evolves as we iteratively update the posterior distributions associated with each hotspot of our test set, for up to ten checklists.  
\paragraph{Species-specific episodic behavior.} We study how our proposed framework can be potentially leveraged to improve the reliability of range maps for individual species in low-data regimes. To do so, we consider how the absolute error between the predicted and ground truth encounter rates for a single species changes as we iteratively update the posterior distribution across an entire region.
\paragraph{Metrics.}
We evaluate our methods with the same metrics as \citet{NEURIPS2023_ef7653bb}, namely MAE, MSE, Top-10, Top-30, and Top-k accuracy. Further details on these metrics are provided in Appendix G. 

\section{Results and Discussion}
\label{section:results}
\paragraph{Overall results.} Table \ref{bigmegatable} details the performance of uncertain-agnostic and uncertainty-aware models on the KE and ZA subdatasets. Results for the US can be found in Tables 10 and 11 in Appendix H.  We observe that uncertainty-aware methods significantly outperforms all the uncertainty-agnostic approaches on our test set, even when prior predictions are updated with five checklists only. This suggests that our Bayesian approach to SDM estimation could greatly improve the reliability of predictions in data-scarce locations. Our proposed framework essentially merges the strengths of traditional tools for Earth-scale mapping, which can make estimates for large regions with high throughput, and of ground-based surveys, which can record specific and accurate data with high effort.

\begin{figure}[h!]
    \centering    \includegraphics[width=\linewidth]{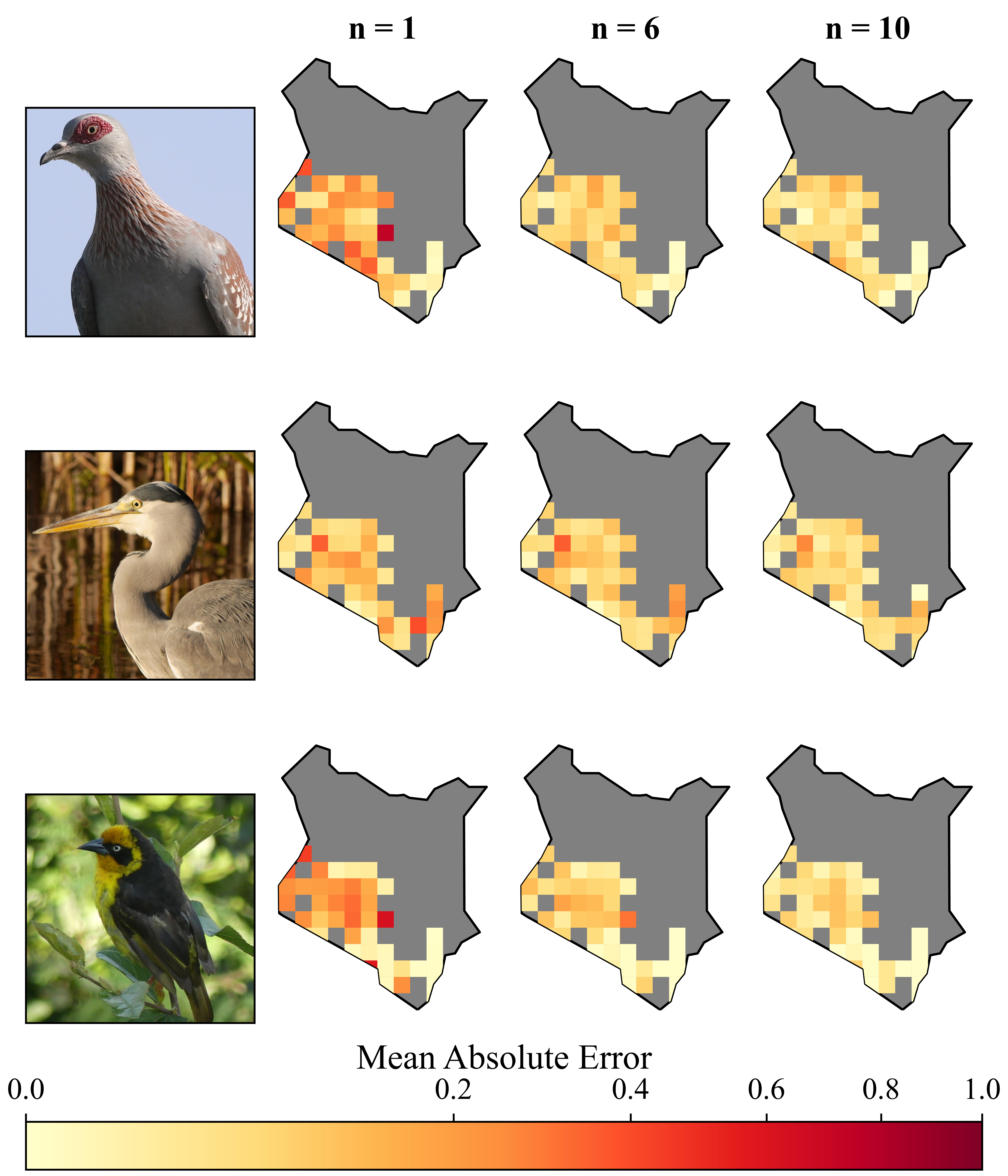}
    \caption{Evolution of the MAE in relation to the number of checklists (1, 6, 10) used to update the posterior distribution for three bird species of Kenya (\textit{Columba guinea}, \textit{Ardea cinerea}, \textit{Ploceus baglafecht}) . The value assigned to each 70km\textsuperscript{2} grid cell corresponds to the mean MAE computed on the aggregation of all the hotspots located within that cell. 
}
    \label{figkenyamapbirds}
\end{figure}

\paragraph{Iterative Improvements by Region.} Figure \ref{figresiterative} shows the evolution of the performance of the uncertainty-aware approaches with increasing number of checklist updates, on the ZA subdataset. Similar figures for the KE and US subdatasets are provided in Appendix H. We observe that our Bayesian framework requires as few as one or two checklists to significantly outperform static baselines. MVN and HetReg are the top-performing methods in our benchmark, with HetReg performing only marginally better than MVN. This is likely because HetReg and MVN share a common aleatoric uncertainty quantification module, but HetReg also includes an epistemic uncertainty estimation module. Estimating epistemic uncertainty alone, however, does not significantly improve on static baselines, as demonstrated by the limited improvements in performance with Deep Ensembles. Our findings are consistent across all four subdatasets. Interestingly, we also observe that the Fixed Variance baseline reaches MAE and MSE performance comparable to HetReg/MVN after 10 checklists, when the prior variance is fixed to its maximum theoretical value.

\paragraph{Iterative Improvements by Species.} As shown in Fig.~\ref{figkenyamapbirds}, our approach rapidly improves encounter rate predictions for species in data-scarce locations. The error quickly drops as more checklists update the posterior distributions, and the improvements are consistent across all hotspots. Our approach shows potential for generating more reliable range maps in areas where a given species has been observed only a handful of times. This could be especially valuable for conservationists monitoring rare or endangered species in areas where gathering additional observations is challenging. 

\paragraph{Aleatoric  vs.~epistemic uncertainty.} We find that models relying on aleatoric uncertainty (such as HetReg and MVN) perform better than those which rely on epistemic uncertainty (such as DE and MCD). This likely results from two factors: First, with real-world observation-based data, similar conditions can occur with very different observations (in this case, reflecting variability in the whereabouts of individual animals and their visibility to an observer). Second, subtle environmental differences may not be fully captured by the input variables. Our conclusions are consistent with trends noted in the Bayesian ML literature \cite{hetregpaper}, in which it is observed that epistemic uncertainty can be poorly estimated in deep learning models due to the implicit assumption that similar inputs lead to similar outcomes. By contrast, aleatoric uncertainty estimation approaches allow us to capture the inherent unpredictability of species observations and the potential existence of local environmental factors not captured by the prior SDM.

\begin{table*}[h!]
\centering
{ 
\fontsize{9}{\baselineskip}\selectfont
\setlength{\tabcolsep}{1mm}
\begin{tabular}{|llllccccccccccl|}
        \cline{4-15}

  \multicolumn{1}{c}{} &    \multicolumn{1}{c}{}                        & \multicolumn{1}{l|}{} &  & \multicolumn{2}{c}{\fontsize{9}{12} MAE[1e-2]} & \multicolumn{2}{c}{MSE[1e-2]} & \multicolumn{2}{c}{Top-10} & \multicolumn{2}{c}{Top-30} & \multicolumn{2}{c}{Top-k} &  \\ \hline 

 \multicolumn{3}{|c}{\cellcolor[HTML]{D3D3D3}}&
\multicolumn{12}{c|}{\cellcolor[HTML]{D3D3D3}\textbf{Kenya}}                                                                                                                                                \\ \hline
  & MER               & \multicolumn{1}{l|}{} &  &           \multicolumn{2}{c|}{\makecell{\colorone{3.94}}} & \multicolumn{2}{c|}{\makecell{\colorone{0.82}}}  &  \multicolumn{2}{c|}{\makecell{\colorone{1.79}}} &   \multicolumn{2}{c|}{\makecell{\colorone{2.63}}}                           &  \multicolumn{2}{c}{\makecell{\colorone{16.21}}}                                      &  \\
  & MLP          & \multicolumn{1}{l|}{} &  &      \multicolumn{2}{c|}{\makecell{\colorone{3.51}\textcolor{gray}{$\pm$0.01}}} & \multicolumn{2}{c|}{\makecell{\colorone{0.81}\textcolor{gray}{$\pm$0.00}}}  &  \multicolumn{2}{c|}{\makecell{\colorone{0.92}\textcolor{gray}{$\pm$0.07}}} &   \multicolumn{2}{c|}{\makecell{\colorone{2.18}\textcolor{gray}{$\pm$0.13}}}                           &  \multicolumn{2}{c}{\makecell{\colorone{16.92}\textcolor{gray}{$\pm$0.04}}}                           &  \\
  & Random Forest                      & \multicolumn{1}{l|}{} &  &     \multicolumn{2}{c|}{\makecell{\colorone{3.57}\textcolor{gray}{$\pm$0.00}}} & \multicolumn{2}{c|}{\makecell{\colorone{1.00}\textcolor{gray}{$\pm$0.00}}}  &  \multicolumn{2}{c|}{\makecell{\colorone{0.31}\textcolor{gray}{$\pm$0.01}}} &   \multicolumn{2}{c|}{\makecell{\colorone{2.05}\textcolor{gray}{$\pm$0.02}}}                           &  \multicolumn{2}{c}{\makecell{\colorone{17.23}\textcolor{gray}{$\pm$0.01}}}                           &  \\
   & ResNet-18                      & \multicolumn{1}{l|}{} &  &     \multicolumn{2}{c|}{\makecell{\colorone{1.87}\textcolor{gray}{$\pm$0.04}}} & \multicolumn{2}{c|}{\makecell{\colorone{0.35}\textcolor{gray}{$\pm$0.01}}}  &  \multicolumn{2}{c|}{\makecell{\colorone{41.24}\textcolor{gray}{$\pm$1.82}}} &   \multicolumn{2}{c|}{\makecell{\colorone{52.93}\textcolor{gray}{$\pm$1.54}}}                           &  \multicolumn{2}{c}{\makecell{\colorone{68.04}\textcolor{gray}{$\pm$0.75}}}                           &  \\\hline\hline
 & ResNet-18+FV                      & \multicolumn{1}{l|}{} &  &     \makecell{\colorone{1.87}\\\textcolor{gray}{$\pm$0.04}} & \multicolumn{1}{c|}{\makecell{\colortwo{1.50}\\\textcolor{gray}{$\pm$0.00}}}  &  \makecell{\colorone{0.35}\\\textcolor{gray}{$\pm$0.01}} &   \multicolumn{1}{c|}{\makecell{\colortwo{0.26}\\\textcolor{gray}{$\pm$0.00}}}                           &  \makecell{\colorone{41.24}\\\textcolor{gray}{$\pm$1.82}}                           &  
  \multicolumn{1}{c|}{\makecell{\colortwo{51.71}\\\textcolor{gray}{$\pm$0.26}}} & \makecell{\colorone{52.93}\\\textcolor{gray}{$\pm$1.54}}  &  \multicolumn{1}{c|}{\makecell{\colortwo{64.46}\\\textcolor{gray}{$\pm$0.31}}} &   \makecell{\colorone{68.82}\\\textcolor{gray}{$\pm$0.75}}                           &  \multicolumn{1}{c}{\makecell{\colortwo{75.26}\\\textcolor{gray}{$\pm$0.63}}}                           &\\
  & ResNet-18+HV                      & \multicolumn{1}{l|}{} &  &       \makecell{\colorone{1.87}\\\textcolor{gray}{$\pm$0.04}} &
  \multicolumn{1}{c|}{\makecell{\colortwo{1.67}\\\textcolor{gray}{$\pm$0.03}}}                           &
  \makecell{\colorone{0.35}\\\textcolor{gray}{$\pm$0.01}}  & 
  \multicolumn{1}{c|}{\makecell{\colortwo{0.24}\\\textcolor{gray}{$\pm$0.01}}}                           &\makecell{\colorone{41.24}\\\textcolor{gray}{$\pm$1.82}} &
  \multicolumn{1}{c|}{\makecell{\colortwo{54.68}\\\textcolor{gray}{$\pm$0.43}}}                           &\makecell{\colorone{52.93}\\\textcolor{gray}{$\pm$1.54}}                           &
  \multicolumn{1}{c|}{\makecell{\colortwo{63.03}\\\textcolor{gray}{$\pm$0.63}}}                           &\makecell{\colorone{68.82}\\\textcolor{gray}{$\pm$0.75}}                           &
  \multicolumn{1}{c}{\makecell{\colortwo{65.82}\\\textcolor{gray}{$\pm$0.70}}}                           &\\
 & ResNet-18+MVN                      & \multicolumn{1}{l|}{} &  &     \makecell{\colorone{2.04}\\\textcolor{gray}{$\pm$0.11}} & \multicolumn{1}{c|}{\makecell{\colortwo{1.52}\\\textcolor{gray}{$\pm$0.03}}}  &  \makecell{\colorone{0.37}\\\textcolor{gray}{$\pm$0.01}} &   \multicolumn{1}{c|}{\makecell{\colortwo{0.24}\\\textcolor{gray}{$\pm$0.00}}}                           &  \makecell{\colorone{36.33}\\\textcolor{gray}{$\pm$1.07}}                           &  
  \multicolumn{1}{c|}{\makecell{\colortwo{57.98}\\\textcolor{gray}{$\pm$0.10}}} & \makecell{\colorone{48.90}\\\textcolor{gray}{$\pm$1.23}}  &  \multicolumn{1}{c|}{\makecell{\colortwo{67.97}\\\textcolor{gray}{$\pm$0.17}}} &   \makecell{\colorone{66.38}\\\textcolor{gray}{$\pm$0.82}}                           &  \multicolumn{1}{c}{\makecell{\colortwo{61.46}\\\textcolor{gray}{$\pm$2.45}}}                           &\\
 & ResNet-18+HetReg                        & \multicolumn{1}{l|}{} &  &        \makecell{\colorone{2.04}\\\textcolor{gray}{$\pm$0.02}} & \multicolumn{1}{c|}{\makecell{\colortwo{1.52}\\\textcolor{gray}{$\pm$0.01}}}  &\makecell{\colorone{0.37}\\\textcolor{gray}{$\pm$0.01}}  &  \multicolumn{1}{c|}{\makecell{\colortwo{0.25}\\\textcolor{gray}{$\pm$0.00}}}  &\makecell{\colorone{36.24}\\\textcolor{gray}{$\pm$1.06}} &\multicolumn{1}{c|}{\makecell{\colortwo{57.98}\\\textcolor{gray}{$\pm$0.13}}}  &   \makecell{\colorone{49.11}\\\textcolor{gray}{$\pm$0.53}}                           & \multicolumn{1}{c|}{\makecell{\colortwo{67.82}\\\textcolor{gray}{$\pm$0.09}}}  & \makecell{\colorone{66.67}\\\textcolor{gray}{$\pm$0.12}}                           &  \multicolumn{1}{c}{\makecell{\colortwo{61.19}\\\textcolor{gray}{$\pm$0.60}}}                           &\\
 & ResNet-18+DE                        & \multicolumn{1}{l|}{} &  &     \makecell{\colorone{1.81}\\\textcolor{gray}{$\pm$0.04}} & \multicolumn{1}{c|}{\makecell{\colortwo{1.75}\\\textcolor{gray}{$\pm$0.06}}}  &  \makecell{\colorone{0.32}\\\textcolor{gray}{$\pm$0.01}} &   \multicolumn{1}{c|}{\makecell{\colortwo{0.29}\\\textcolor{gray}{$\pm$0.01}}}                           &  \makecell{\colorone{44.34}\\\textcolor{gray}{$\pm$0.31}}                           &  
  \multicolumn{1}{c|}{\makecell{\colortwo{48.47}\\\textcolor{gray}{$\pm$1.06}}} & \makecell{\colorone{56.06}\\\textcolor{gray}{$\pm$0.63}}  &  \multicolumn{1}{c|}{\makecell{\colortwo{59.54}\\\textcolor{gray}{$\pm$1.48}}} &   \makecell{\colorone{69.94}\\\textcolor{gray}{$\pm$0.53}}                           &  \multicolumn{1}{c}{\makecell{\colortwo{70.41}\\\textcolor{gray}{$\pm$0.56}}}                           &\\
  & ResNet-18+MCD                       & \multicolumn{1}{l|}{} &  &     \makecell{\colorone{1.91}\\\textcolor{gray}{$\pm$0.01}} & \multicolumn{1}{c|}{\makecell{\colortwo{1.82}\\\textcolor{gray}{$\pm$0.01}}}  &  \makecell{\colorone{0.37}\\\textcolor{gray}{$\pm$0.01}} &   \multicolumn{1}{c|}{\makecell{\colortwo{0.33}\\\textcolor{gray}{$\pm$0.01}}}                           &  \makecell{\colorone{38.65}\\\textcolor{gray}{$\pm$1.19}}                           &  
  \multicolumn{1}{c|}{\makecell{\colortwo{43.15}\\\textcolor{gray}{$\pm$1.01}}} & \makecell{\colorone{50.18}\\\textcolor{gray}{$\pm$0.63}}  &  \multicolumn{1}{c|}{\makecell{\colortwo{54.36}\\\textcolor{gray}{$\pm$0.81}}} &   \makecell{\colorone{67.18}\\\textcolor{gray}{$\pm$0.11}}                           &  \multicolumn{1}{c}{\makecell{\colortwo{68.30}\\\textcolor{gray}{$\pm$0.19}}}                           &\\
   & ResNet-18+SE                       & \multicolumn{1}{l|}{} &  &     \makecell{\colorone{1.73}\\\textcolor{gray}{$\pm$0.01}} & \multicolumn{1}{c|}{\makecell{\colortwo{1.71}\\\textcolor{gray}{$\pm$0.01}}}  &  \makecell{\colorone{0.33}\\\textcolor{gray}{$\pm$0.00}} &   \multicolumn{1}{c|}{\makecell{\colortwo{0.32}\\\textcolor{gray}{$\pm$0.00}}}                           &  \makecell{\colorone{43.53}\\\textcolor{gray}{$\pm$0.26}}                           &  
  \multicolumn{1}{c|}{\makecell{\colortwo{44.19}\\\textcolor{gray}{$\pm$0.08}}} & \makecell{\colorone{55.79}\\\textcolor{gray}{$\pm$0.41}}  &  \multicolumn{1}{c|}{\makecell{\colortwo{56.49}\\\textcolor{gray}{$\pm$0.44}}} &   \makecell{\colorone{70.47}\\\textcolor{gray}{$\pm$0.16}}                           &  \multicolumn{1}{c}{\makecell{\colortwo{70.70}\\\textcolor{gray}{$\pm$0.16}}}                           &\\
  \hline
 
 \multicolumn{3}{|c}{\cellcolor[HTML]{D3D3D3}}&
\multicolumn{12}{c|}{\cellcolor[HTML]{D3D3D3}\textbf{South Africa}}                          
\\ \hline
  & MER              & \multicolumn{1}{l|}{} &  &         \multicolumn{2}{c|}{\makecell{\colorone{3.62}}} & \multicolumn{2}{c|}{\makecell{\colorone{0.84}}}  &  \multicolumn{2}{c|}{\makecell{\colorone{25.53}}} &   \multicolumn{2}{c|}{\makecell{\colorone{36.87}}}                           &  \multicolumn{2}{c}{\makecell{\colorone{49.31}}}                           &  \\
  & MLP         & \multicolumn{1}{l|}{} &  &         \multicolumn{2}{c|}{\makecell{\colorone{2.87}\textcolor{gray}{$\pm$0.00}}} & \multicolumn{2}{c|}{\makecell{\colorone{0.71}\textcolor{gray}{$\pm$0.00}}}  &  \multicolumn{2}{c|}{\makecell{\colorone{35.17}\textcolor{gray}{$\pm$0.15}}} &   \multicolumn{2}{c|}{\makecell{\colorone{51.18}\textcolor{gray}{$\pm$0.4}}}                           &  \multicolumn{2}{c}{\makecell{\colorone{61.72}\textcolor{gray}{$\pm$0.10}}}                           &  \\
  & Random Forest                     & \multicolumn{1}{l|}{} &  &         \multicolumn{2}{c|}{\makecell{\colorone{2.54}\textcolor{gray}{$\pm$0.01}}} & \multicolumn{2}{c|}{\makecell{\colorone{0.69}\textcolor{gray}{$\pm$0.00}}}  &  \multicolumn{2}{c|}{\makecell{\colorone{38.21}\textcolor{gray}{$\pm$0.16}}} &   \multicolumn{2}{c|}{\makecell{\colorone{54.23}\textcolor{gray}{$\pm$0.20}}}                           &  \multicolumn{2}{c}{\makecell{\colorone{64.26}\textcolor{gray}{$\pm$0.09}}}                           &  \\
  & ResNet-18                          & \multicolumn{1}{l|}{} &  &     \multicolumn{2}{c|}{\makecell{\colorone{2.47}\textcolor{gray}{$\pm$0.02}}} & \multicolumn{2}{c|}{\makecell{\colorone{0.62}\textcolor{gray}{$\pm$0.01}}}  &  \multicolumn{2}{c|}{\makecell{\colorone{41.79}\textcolor{gray}{$\pm$0.59}}} &   \multicolumn{2}{c|}{\makecell{\colorone{57.20}\textcolor{gray}{$\pm$0.32}}}                           &  \multicolumn{2}{c}{\makecell{\colorone{67.11}\textcolor{gray}{$\pm$0.23}}}                           &  \\ \hline \hline
  & ResNet-18+FV                      & \multicolumn{1}{l|}{} &  &     \makecell{\colorone{2.47}\\\textcolor{gray}{$\pm$0.02}} & \multicolumn{1}{c|}{\makecell{\colortwo{1.65}\\\textcolor{gray}{$\pm$0.00}}}  &  \makecell{\colorone{0.62}\\\textcolor{gray}{$\pm$0.01}} &   \multicolumn{1}{c|}{\makecell{\colortwo{0.35}\\\textcolor{gray}{$\pm$0.00}}}                           &  \makecell{\colorone{41.79}\\\textcolor{gray}{$\pm$0.59}}                           &  
  \multicolumn{1}{c|}{\makecell{\colortwo{57.00}\\\textcolor{gray}{$\pm$0.17}}} & \makecell{\colorone{57.20}\\\textcolor{gray}{$\pm$0.32}}  &  \multicolumn{1}{c|}{\makecell{\colortwo{69.59}\\\textcolor{gray}{$\pm$0.12}}} &   \makecell{\colorone{67.11}\\\textcolor{gray}{$\pm$0.23}}                           &  \multicolumn{1}{c}{\makecell{\colortwo{79.90}\\\textcolor{gray}{$\pm$0.13}}}                           &\\
  & ResNet-18+HV                      & \multicolumn{1}{l|}{} &  &       \makecell{\colorone{2.47}\\\textcolor{gray}{$\pm$0.04}} &
  \multicolumn{1}{c|}{\makecell{\colortwo{2.06}\\\textcolor{gray}{$\pm$0.02}}}                           &
  \makecell{\colorone{0.62}\\\textcolor{gray}{$\pm$0.00}}  & 
  \multicolumn{1}{c|}{\makecell{\colortwo{0.34}\\\textcolor{gray}{$\pm$0.00}}}                           &\makecell{\colorone{41.79}\\\textcolor{gray}{$\pm$0.59}} &
  \multicolumn{1}{c|}{\makecell{\colortwo{56.00}\\\textcolor{gray}{$\pm$0.32}}}                           &\makecell{\colorone{57.20}\\\textcolor{gray}{$\pm$0.32}}                           &
  \multicolumn{1}{c|}{\makecell{\colortwo{67.55}\\\textcolor{gray}{$\pm$0.23}}}                           &\makecell{\colorone{67.11}\\\textcolor{gray}{$\pm$0.23}}                           &
  \multicolumn{1}{c}{\makecell{\colortwo{70.79}\\\textcolor{gray}{$\pm$0.13}}}                           &\\
  & ResNet-18+MVN                      & \multicolumn{1}{l|}{} &  &     \makecell{\colorone{2.66}\\\textcolor{gray}{$\pm$0.02}} & \multicolumn{1}{c|}{\makecell{\colortwo{1.66}\\\textcolor{gray}{$\pm$0.00}}}  &  \makecell{\colorone{0.65}\\\textcolor{gray}{$\pm$0.01}} &   \multicolumn{1}{c|}{\makecell{\colortwo{0.32}\\\textcolor{gray}{$\pm$0.00}}}                           &  \makecell{\colorone{37.01}\\\textcolor{gray}{$\pm$1.12}}                           &  
  \multicolumn{1}{c|}{\makecell{\colortwo{60.19}\\\textcolor{gray}{$\pm$0.16}}} & \makecell{\colorone{53.20}\\\textcolor{gray}{$\pm$1.06}}  &  \multicolumn{1}{c|}{\makecell{\colortwo{70.68}\\\textcolor{gray}{$\pm$0.17}}} &   \makecell{\colorone{64.61}\\\textcolor{gray}{$\pm$0.27}}                           &  \multicolumn{1}{c}{\makecell{\colortwo{71.88}\\\textcolor{gray}{$\pm$0.35}}}                           &\\
  & ResNet-18+HetReg                      & \multicolumn{1}{l|}{} &  &       \makecell{\colorone{2.70}\\\textcolor{gray}{$\pm$0.02}} &
  \multicolumn{1}{c|}{\makecell{\colortwo{1.67}\\\textcolor{gray}{$\pm$0.00}}}                           &
  \makecell{\colorone{0.64}\\\textcolor{gray}{$\pm$0.00}}  & 
  \multicolumn{1}{c|}{\makecell{\colortwo{0.32}\\\textcolor{gray}{$\pm$0.00}}}                           &\makecell{\colorone{37.63}\\\textcolor{gray}{$\pm$0.58}} &
  \multicolumn{1}{c|}{\makecell{\colortwo{60.10}\\\textcolor{gray}{$\pm$0.03}}}                           &\makecell{\colorone{53.65}\\\textcolor{gray}{$\pm$0.18}}                           &
  \multicolumn{1}{c|}{\makecell{\colortwo{70.5}\\\textcolor{gray}{$\pm$0.04}}}                           &\makecell{\colorone{64.84}\\\textcolor{gray}{$\pm$0.17}}                           &
  \multicolumn{1}{c}{\makecell{\colortwo{71.0}\\\textcolor{gray}{$\pm$0.30}}}                           &\\
  & ResNet-18+DE                       & \multicolumn{1}{l|}{} &  &     \makecell{\colorone{2.44}\\\textcolor{gray}{$\pm$0.01}} & \multicolumn{1}{c|}{\makecell{\colortwo{2.31}\\\textcolor{gray}{$\pm$0.01}}}  &  \makecell{\colorone{0.60}\\\textcolor{gray}{$\pm$0.00}} &   \multicolumn{1}{c|}{\makecell{\colortwo{0.53}\\\textcolor{gray}{$\pm$0.00}}}                           &  \makecell{\colorone{43.30}\\\textcolor{gray}{$\pm$0.13}}                           &  
  \multicolumn{1}{c|}{\makecell{\colortwo{48.06}\\\textcolor{gray}{$\pm$0.12}}} & \makecell{\colorone{58.47}\\\textcolor{gray}{$\pm$0.06}}  &  \multicolumn{1}{c|}{\makecell{\colortwo{62.11}\\\textcolor{gray}{$\pm$0.22}}} &   \makecell{\colorone{67.84}\\\textcolor{gray}{$\pm$0.13}}                           &  \multicolumn{1}{c}{\makecell{\colortwo{69.72}\\\textcolor{gray}{$\pm$0.09}}}                           &\\
 & ResNet-18+MCD                       & \multicolumn{1}{l|}{} &  &     \makecell{\colorone{2.61}\\\textcolor{gray}{$\pm$0.01}} & \multicolumn{1}{c|}{\makecell{\colortwo{2.47}\\\textcolor{gray}{$\pm$0.01}}}  &  \makecell{\colorone{0.64}\\\textcolor{gray}{$\pm$0.00}} &   \multicolumn{1}{c|}{\makecell{\colortwo{0.56}\\\textcolor{gray}{$\pm$0.00}}}                           &  \makecell{\colorone{39.21}\\\textcolor{gray}{$\pm$0.44}}                           &  
  \multicolumn{1}{c|}{\makecell{\colortwo{45.21}\\\textcolor{gray}{$\pm$0.30}}} & \makecell{\colorone{55.00}\\\textcolor{gray}{$\pm$0.40}}  &  \multicolumn{1}{c|}{\makecell{\colortwo{59.22}\\\textcolor{gray}{$\pm$0.35}}} &   \makecell{\colorone{65.64}\\\textcolor{gray}{$\pm$0.16}}                           &  \multicolumn{1}{c}{\makecell{\colortwo{67.86}\\\textcolor{gray}{$\pm$0.20}}}                           &\\
   & ResNet-18+SE                       & \multicolumn{1}{l|}{} &  &     \makecell{\colorone{2.48}\\\textcolor{gray}{$\pm$0.03}} & \multicolumn{1}{c|}{\makecell{\colortwo{2.46}\\\textcolor{gray}{$\pm$0.02}}}  &  \makecell{\colorone{0.63}\\\textcolor{gray}{$\pm$0.00}} &   \multicolumn{1}{c|}{\makecell{\colortwo{0.61}\\\textcolor{gray}{$\pm$0.00}}}                           &  \makecell{\colorone{41.90}\\\textcolor{gray}{$\pm$0.22}}                           &  
  \multicolumn{1}{c|}{\makecell{\colortwo{42.80}\\\textcolor{gray}{$\pm$0.39}}} & \makecell{\colorone{56.93}\\\textcolor{gray}{$\pm$0.10}}  &  \multicolumn{1}{c|}{\makecell{\colortwo{57.67}\\\textcolor{gray}{$\pm$0.13}}} &   \makecell{\colorone{67.09}\\\textcolor{gray}{$\pm$0.09}}                           &  \multicolumn{1}{c}{\makecell{\colortwo{67.48}\\\textcolor{gray}{$\pm$0.06}}}                           &\\
  \hline    \toprule

\end{tabular}
}
\caption{Performance of SDM-estimation approaches both \colorone{without Bayesian updates} (left side in split columns) and \colortwo{after updates from five checklists} (right side). Note that the first four methods for each region are not uncertainty-aware and therefore cannot be updated with checklist information. Means are shown $\pm$ standard deviation across runs. (Note that many standard deviations are quite low; those reported as 0.00 are accurate.)}
\label{bigmegatable}
\end{table*}

\section{Limitations}
\paragraph{Effectively combining large-scale ecological patterns with local information remains challenging.} There remains tension between the broad environmental variables learned by deep SDMs (our Bayesian prior) and the hotspot-specific features captured by observational information (our updates to that prior). As shown in Figure \ref{figresiterative}, the initial predictions of an uncertainty-aware SDM can be overwritten in a first Bayesian update, when the estimated prior variance is large. This can lead to an initial drop in performance before enough updates are performed to boost performance, even though performance would ideally increase \textit{monotonically} with number of updates. A simple approach to mitigate such behavior is to consider a weighted average of predictions from the prior and the updated SDM, progressively giving more importance to the updates with more observations. We provide additional experiments showing how such an approach can help smooth the results curve observed in Fig. \ref{figresiterative} in Appendix E..

\paragraph{Uncertainty estimation approaches cannot fully correct for biases in citizen science data.} Citizen science data is heavily influenced by \textit{predictable} 
factors such as the day of the week, the time of day, the expertise of the observer, the weather, and the proximity to population centers \cite{sierra2025divshift}. Such biases are \textit{structured}, which technically violates the randomness assumption of aleatoric uncertainty. Epistemic uncertainty estimation approaches can potentially correct for biases if they can be explicitly modeled, but are unable to do so if they are not encoded in the data. While our work shows that is possible to compensate partially for data gaps associated with the irregular distribution of citizen science data, some biases may affect the accuracy of the underlying numbers used as ground truth, which cannot be compensated for by such methods and remain an active object of inquiry in statistical ecology \cite{spatialbiasref}.

\section{Conclusion}
In this work, {we presented BATIS, a novel framework for species distribution modeling in which uncertainty-aware machine learning approaches are iteratively refined with additional on-the-ground information. Our results show how our framework can rapidly improve SDM predictions in data-scarce locations by combining
broader ecological patterns with fine-grained, location-specific insights.  Our study also suggests that ML approaches focusing on aleatoric uncertainty provide a better measure of observational variability than those focusing on epistemic uncertainty.
\paragraph{Pathway to impact.}

SDMs are widely used to inform decisions in conservation, land use, and other areas of impact \cite{sdmessentialtool}. We hope that BATIS can improve the accuracy of such SDMs in applications where new observations are continually arriving, such as the monitoring of protected areas and endangered species, as well as guiding policies in underresourced or remote regions with lower data availability. Our framework is designed to be readily integrated into existing workflows for creation and utilization of SDMs. In addition, traditional barriers to integration of deep learning into ecology have included computational cost and lack of perceived interpretability. Our proposed approach is computationally lightweight and  allow for improved performance without costly retraining, while the focus on uncertainty quantification makes model predictions more transparent, as well as bringing the output closer to the statistical tools already familiar to those working with SDMs.

\paragraph{Future work.} Future work includes extending BATIS to guide sampling effort in citizen science initiatives, in collaboration with ecologists, and investigating the behavior of different uncertainty estimation approaches with presence-only data (which are often more abundant and readily available than presence-absence checklist data ; \citet{zbindenpresenceabsence}). Finally, an impactful extension of our work would be to control for additional sources of biases in the data, e.g. building on \citet{sierra2025divshift} by considering subdatasets labeled according to different types of bias.

\section*{Acknowledgments}
We would like to thank members of the Rolnick lab and Lauren Harrell for feedback on earlier versions of this work and for helpful discussions. This work was partly supported by the Global Center on AI and Biodiversity Change (US NSF OISE-2330423 and Canada NSERC 585136) and the Canada CIFAR AI Chairs program. Catherine Villeneuve is also supported by an NSERC CGS-D fellowship (Award No. 589535). We also acknowledge computational support from the Digital Research Alliance of Canada and from Mila - AI Quebec AI Institute, including material support from NVIDIA Corporation. 

\bibliography{aaai2026}

\clearpage 
\appendix

\section{Dataset}
\label{appendix:dataset-splits}
\subsection{Input Variables}
\paragraph{Satellite Imagery.} We used remote sensing imagery from the Sentinel-2 satellite.  Sentinel-2 data is publicly available and can be easily extracted with Google EarthEngine, using their \texttt{COPERNICUS/S2\_SR\_HARMONIZED} dataset from their Data Catalog. We provide code to extract Sentinel-2 data from EarthEngine in our Github repository. A summary of the bands we extracted (RGBNIR) can be found in Table \ref{tab:bandsforsentinel2}. Each band has a resolution of 10m/pixel. We strictly considered images with a cloud coverage of up to  10\%. 
\begin{table}[h!]
\centering
{
\fontsize{9}{\baselineskip}\selectfont
\begin{tabular}{|lll|}
\hline
\multicolumn{3}{|c|}{\cellcolor[HTML]{C0C0C0} \textbf{Sentinel-2 Bands}}                                       \\ \hline
\multicolumn{1}{|l|}{\textbf{Band}} & \multicolumn{1}{l|}{\textbf{Description}} & \textbf{Resolution} \\ \hline
\multicolumn{1}{|l|}{B4}            & \multicolumn{1}{l|}{Red}                  & 10m/pixel           \\ \hline
\multicolumn{1}{|l|}{B3}            & \multicolumn{1}{l|}{Green}                & 10m/pixel           \\ \hline
\multicolumn{1}{|l|}{B2}            & \multicolumn{1}{l|}{Blue}                 & 10m/pixel           \\ \hline
\multicolumn{1}{|l|}{B8}            & \multicolumn{1}{l|}{Near-Infrared (NIR)}        & 10m/pixel           \\ \hline
\end{tabular}
}
\caption{Description of the four bands we extracted from Sentinel-2 rasters}
\label{tab:bandsforsentinel2}
\end{table}

\paragraph{Bioclimatic Variables.} We used the 19 bioclimatic variables of the WorldClim model \citet{worldclim}. For the USA-Summer and USA-Winter subsets, we followed \citet{NEURIPS2023_ef7653bb, geolifeclef} and considered data from WorldClim 1.4, which has data aggregated
from 1960-1990. For Kenya and South Africa, we considered data from WorldClim 2.1 instead, which has data aggregated from 1970-2000. Both WorldClim 1.4 and WorldClim 2.1 share the same variables and the same resolution (approximately 1km per pixel). Table \ref{tab:bioclimvariables} summarizes the variables we used. 
\begin{table*}[h!]
\centering
{
\fontsize{9}{\baselineskip}\selectfont
\begin{tabular}{|lll|}
\Xhline{1.2pt}
\multicolumn{3}{|c|}{\cellcolor[HTML]{C0C0C0}\textbf{Bioclimatic Variables}}                                                                            \\ \Xhline{1.2pt} 
\multicolumn{1}{|l|}{\textbf{Name}} & \multicolumn{1}{l|}{\textbf{Description}}                                       & \textbf{Unit}            \\ \Xhline{1pt}
\multicolumn{1}{|l|}{\texttt{bio\_1}}        & \multicolumn{1}{l|}{Annual Mean Temperature}                                    & °C                       \\ \hline
\multicolumn{1}{|l|}{\texttt{bio\_2}}    & 
\multicolumn{1}{l|}{Mean Diurnal Range (Mean of monthly (max temp - min temp))} & °C                       \\ \hline
\multicolumn{1}{|l|}{\texttt{bio\_3}}        & \multicolumn{1}{l|}{Isothermality (BIO2/BIO7) (×100)}                           & \%                       \\ \hline
\multicolumn{1}{|l|}{\texttt{bio\_4}}      & \multicolumn{1}{l|}{Temperature Seasonality (standard deviation ×100)}          & °C                       \\ \hline
\multicolumn{1}{|l|}{\texttt{bio\_5}}        & \multicolumn{1}{l|}{Max Temperature of Warmest Month}                           & °C                       \\ \hline
\multicolumn{1}{|l|}{\texttt{bio\_6}}     & \multicolumn{1}{l|}{Min Temperature of Coldest Month}                           & °C                       \\ \hline
\multicolumn{1}{|l|}{\texttt{bio\_7}}        & \multicolumn{1}{l|}{Temperature Annual Range (BIO5-BIO6)}                       & °C                       \\ \hline
\multicolumn{1}{|l|}{\texttt{bio\_8}}        & \multicolumn{1}{l|}{Mean Temperature of Wettest Quarter}                        & °C                       \\ \hline
\multicolumn{1}{|l|}{\texttt{bio\_9}}        & \multicolumn{1}{l|}{Mean Temperature of Driest Quarter}                         & °C                       \\ \hline
\multicolumn{1}{|l|}{\texttt{bio\_10}}   & \multicolumn{1}{l|}{Mean Temperature of Warmest Quarter}                        & °C                       \\ \hline
\multicolumn{1}{|l|}{\texttt{bio\_11}}      & \multicolumn{1}{l|}{Mean Temperature of Coldest Quarter}                        & °C                       \\ \hline
\multicolumn{1}{|l|}{\texttt{bio\_12}}       & \multicolumn{1}{l|}{Annual Precipitation}                                       & mm                       \\ \hline
\multicolumn{1}{|l|}{\texttt{bio\_13}}      & \multicolumn{1}{l|}{Precipitation of Wettest Month}                             & mm                       \\ \hline
\multicolumn{1}{|l|}{\texttt{bio\_14}}     & \multicolumn{1}{l|}{Precipitation of Driest Month}                              & mm                       \\ \hline
\multicolumn{1}{|l|}{\texttt{bio\_15}}      & \multicolumn{1}{l|}{Precipitation Seasonality}                                  & Coefficient of variation \\ \hline
\multicolumn{1}{|l|}{\texttt{bio\_16}}     & \multicolumn{1}{l|}{Precipitation of Wettest Quarter}                           & mm                       \\ \hline
\multicolumn{1}{|l|}{\texttt{bio\_17}}     & \multicolumn{1}{l|}{Precipitation of Driest Quarter}                            & mm                       \\ \hline
\multicolumn{1}{|l|}{\texttt{bio\_18}}  & \multicolumn{1}{l|}{Precipitation of Warmest Quarter}                           & mm                       \\ \hline
\multicolumn{1}{|l|}{\texttt{bio\_19}}     & \multicolumn{1}{l|}{Precipitation of Coldest Quarter}                           & mm                       \\ \Xhline{1.2pt}
\end{tabular}%
}
\caption{Description of the bioclimatic variables extracted from WorldClim. Each of these rasters has a resolution of approximately 1km per pixel}
\label{tab:bioclimvariables}
\end{table*}

\paragraph{MLP/RF Inputs.} The pipeline we leveraged to build model inputs for our MLP and Random Forest baselines is summarized by Algorithm \ref{alg:inputsmlprf}. 

\begin{algorithm}[h!]
\caption{Input Generation Pipeline for MLP and Random Forest baselines}
\label{alg:inputsmlprf}
\textbf{Input :} List of hotspots $\mathcal{H} = \{h_1, h_2, ..., h_n\}$ \\ 
\textbf{Output :} Processed model inputs for each hotspot
\begin{algorithmic}
\FOR{each $h_{k}$ in $\mathcal{H}$}
   \STATE \textbf{Step 1 :} Retrieve  \texttt{(lat, lon)} coordinates associated with $h_{k}$ from the eBird DataBase
 \STATE \textbf{Step 2 :} From \texttt{(lat, lon)}, retrieve an input vector $\mathbf{x_{k}} \in \mathbb{R}^{19}$ from WorldClim rasters
\ENDFOR
\end{algorithmic}
\end{algorithm}

\paragraph{Resnet-18 Inputs.} The pipeline we leveraged to build model inputs for our Resnet18-based models is summarized by Algorithm \ref{alg:inputsresnet18}. Note that the conversion to latitude/longitude coordinates to UTM projection system is a standard practice, as it allows Euclidean distances to be computed in meters. It therefore becomes straightforward to crop a region of a pre-determined size from any raster. 
\begin{algorithm}[h!]
\caption{Input Generation Pipeline for Resnet18-based models}
\label{alg:inputsresnet18}
\textbf{Input :} List of hotspots $\mathcal{H} = \{h_1, h_2, ..., h_n\}$ \\
\textbf{Output :} Processed model inputs for each hotspot
\begin{algorithmic}
\FOR{each $h_{k}$ in $\mathcal{H}$}
    \STATE \textbf{Step 1 :} Retrieve  \texttt{(lat, lon)} coordinates associated with $h_{k}$ from the eBird DataBase

   \STATE \textbf{Step 2 :} Convert \texttt{(lat, lon)} coordinates into UTM projection \texttt{(utm\_e, utm\_n)}
   
    \STATE \textbf{Step 3 :} Build a 5km\textsuperscript{2} bounding box centered on \texttt{(utm\_e, utm\_n)} 
    
    \STATE \textbf{Step 4 :} Use the bouding box obtained from previous step to extract \texttt{sat\_img} raster from Sentinel-2 and \texttt{env\_var} raster from WorldClim  
    
    \STATE \textbf{Step 5 :} Use bilinear interpolation to match \texttt{env\_var} (1km/pixel) resolution to \texttt{sat\_img} resolution (10m/pixel)

    \STATE \textbf{Step 6 :} From each \texttt{(sat\_img, env\_var)} pair, retrieve an input matrix $\mathbf{X}_{k} \in \mathbb{R}^{64 \times 64 \times 19}$ by center-cropping a region of 640m\textsuperscript{2}    
\ENDFOR
\end{algorithmic}
\end{algorithm}
\subsection{Splits}
Table \ref{tab:compositionsplits} describes the number of hotspots and the number of checklists associated with the four subsets of our dataset, for each split (train/val/test). These splits were obtained following the process described in section \ref{sec:benchmark}. Similarly to \cite{NEURIPS2023_ef7653bb} and following recommendations in the literature \cite{autocorrgis}, we pre-processed the data based on the geospatial coordinates of each hotspot in order to prevent auto-correlation and over-fitting using DBSCAN to cluster hotspots within 5km of each other and ensuring that individual members of each cluster were proportionally distributed across splits.  

\begin{table}[h!]
\centering
{\fontsize{9}{\baselineskip}\selectfont
\renewcommand{\arraystretch}{1.0}
\begin{tabular}{|l|c|c|c|}
\cline{2-4} \multicolumn{1}{c}{}
& \multicolumn{1}{|c|}{\textbf{Training}}  & \textbf{Validation} & \textbf{Testing} \\
\hline
\multicolumn{4}{|c|}{\cellcolor[HTML]{C0C0C0}\textbf{Kenya}} \\ \hline
\textbf{N. Hotspots}   & 6,481   & 1,852   & 218     \\
\textbf{N. Checklists} & 24,532  & 6,380   & 13,940  \\
\hline
\multicolumn{4}{|c|}{\cellcolor[HTML]{C0C0C0}\textbf{South Africa}} \\ \hline
\textbf{N. Hotspots}   & 5,372   & 672     & 599     \\
\textbf{N. Checklists} & 137,740 & 12,051  & 349,076 \\
\hline
\multicolumn{4}{|c|}{\cellcolor[HTML]{C0C0C0}\textbf{USA-Winter}} \\ \hline
\textbf{N. Hotspots}   & 26,422   & 5,747    & 17,173   \\
\textbf{N. Checklists} & 1,626,861 & 420,020 & 1,941,704 \\
\hline
\multicolumn{4}{|c|}{\cellcolor[HTML]{C0C0C0}\textbf{USA-Summer}} \\ \hline
\textbf{N. Hotspots}   & 63,796   & 14,761   & 24,932   \\
\textbf{N. Checklists} & 1,721,678 & 477,490 & 1,895,054 \\
\hline
\end{tabular}
\caption{Composition of each split for each dataset}
\label{tab:compositionsplits}
}
\end{table}

\subsection{Species List}
\label{appendix:species-list}

Our dataset only considers the regularly occurring species for each region and exclude observations of very rare species (e.g.~Eurasian birds occasionally blown by storms to the US), as such sightings are both unpredictable and do not generally provide any meaningful information about the ecosystem.

\paragraph{Kenya.} We limited the species list to the 1054 species regularly found in Kenya according to Avibase \cite{Avibase}. 

\paragraph{South Africa.} We limited the species list to the 755 species regularly found in South Africa according to Bird Life South Africa \cite{BirdLife}. 

\paragraph{USA-Winter and USA-Summer.} We limited the species list to the 670 species regularly found in continental United States according to the 2022 Checklist of the Continental United States from the American Birding Association \cite{ABA}. More specifically, we only considered species denoted by ABA Codes 1 or 2,
which include regular breeding species and visitors that are widespread across the country. We did not consider the species that are known to only occur in Alaska, Hawaii or the U.S. Territories (e.g. Guam, Puerto Rico, American Samoa, etc.). 

\subsection{Dataset Overview : Kenya}
We provide a visual overview of the Kenya subdataset below. Figure \ref{figsuppkenya} shows the geographical distribution of hotspots and species, as well as the distribution of non-zero encounter rates per hotspot. 

\subsection{Dataset Overview : USA-Winter}

We provide a visual overview of the USA-Winter subdataset below. Figure \ref{figsuppwinter} shows the geographical distribution of hotspots and species, as well as the distribution of non-zero encounter rates per hotspot. 

\subsection{Dataset Overview : South Africa}

We provide a visual overview of the South-Africa subdataset below. Figure \ref{figsuppza} shows the geographical distribution of hotspots and species, as well as the distribution of non-zero encounter rates per hotspot.

\subsection{Dataset Overview : USA-Summer}

We provide a visual overview of the USA-Winter subdataset below. Figure \ref{figsuppsummer} shows the geographical distribution of hotspots and species, as well as the distribution of non-zero encounter rates per hotspot. 

\section{HuggingFace Repository}
\label{appendix:huggingfacerepo} 

Our Dataset is available on HuggingFace. More information can be found in the \texttt{README.md} of the repository. 

\section{Code Repository}
\label{appendix:githubrepository} 

Our Code is available on GitHub. More information can be found in the \texttt{README.md} of the repository. 

\section{Licenses}
\label{appendix:licenses} 
The BATIS Benchmark is released under a Creative Commons Attribution-NonCommercial
4.0 International (CC BY-NC 4.0) License. The use of our dataset should comply with the eBird terms of Use, the eBird API Terms of use, and the eBird Data Access Terms of Use.

\begin{figure*}[htbp]
    \centering
    \includegraphics[width=\linewidth]{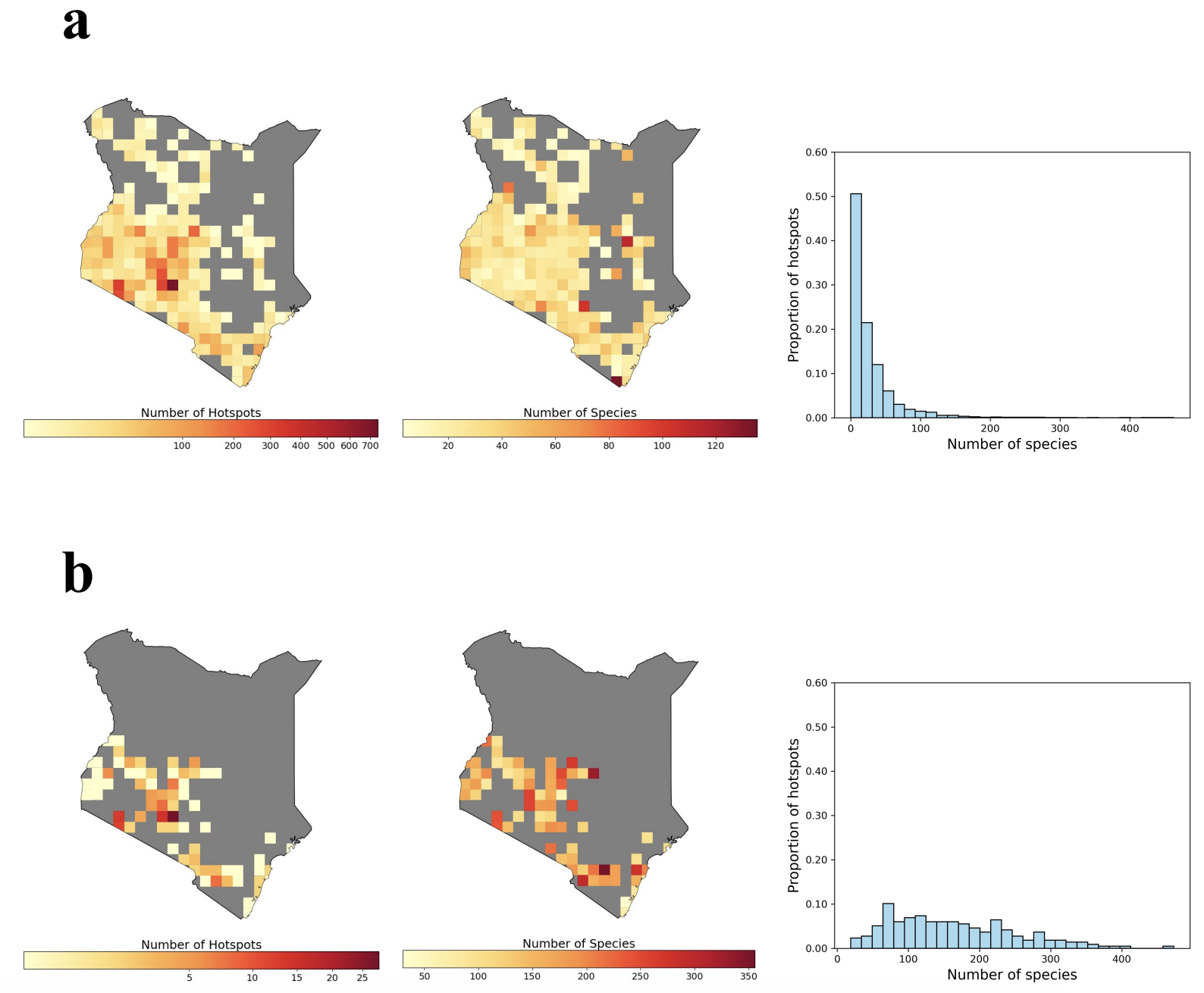}
    \caption{From left to right, for a) Training and b) Test : Geographical distribution of the total number of hotspots (left) and non-zero encounter rates (middle) associated with our Kenya subdataset, and distribution of the number of species encountered per hotspot (right).}
    \label{figsuppkenya}
\end{figure*}

\begin{figure*}[htbp]
    \centering
    \includegraphics[width=\linewidth]{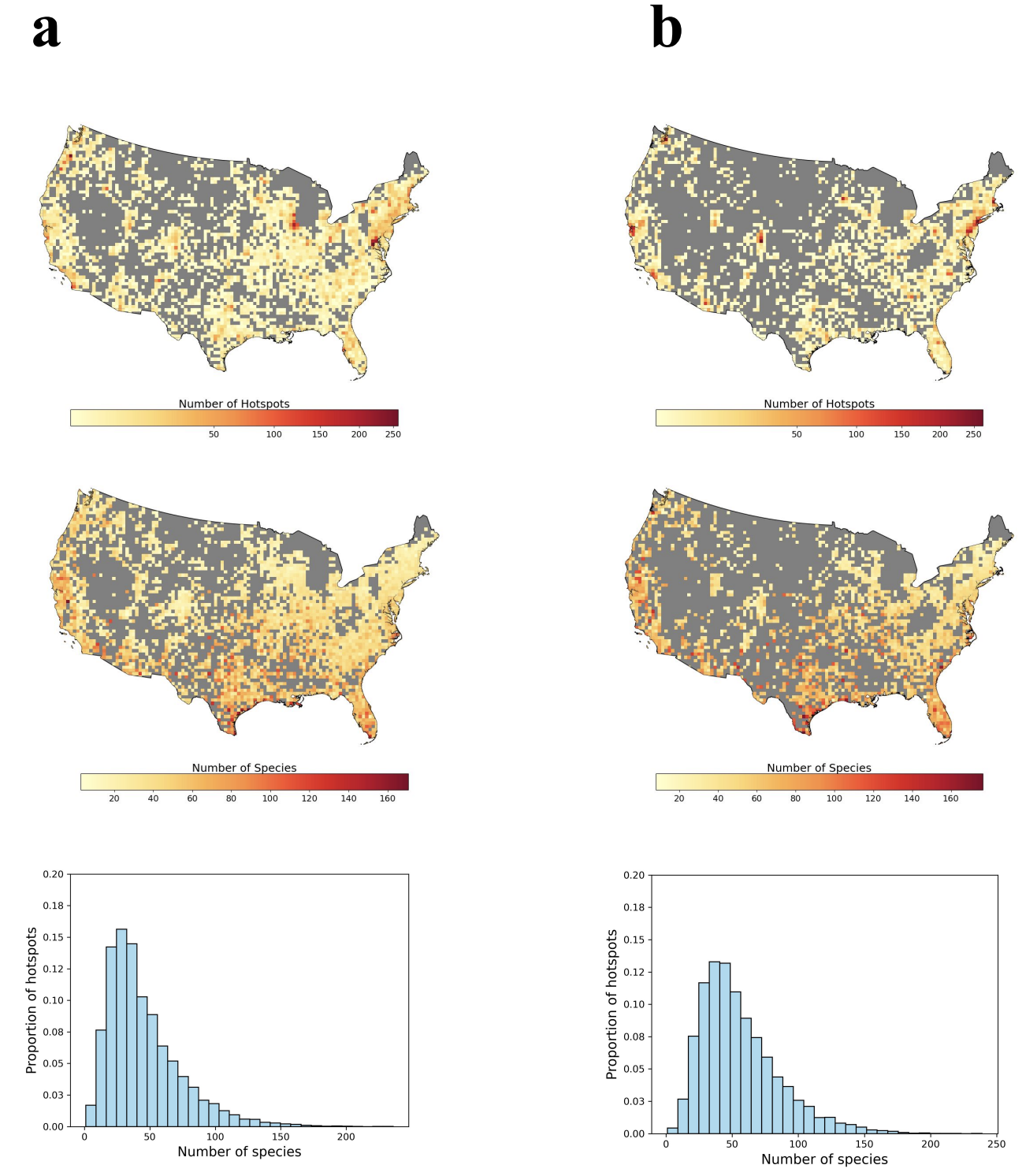}
    \caption{From top to bottom, for a) Training and b) Test : Geographical distribution of the total number of hotspots (top) and non-zero encounter rates (middle) associated with our US-Winter subdataset, and distribution of the number of species encountered per hotspot (bottom).}
    \label{figsuppwinter}
\end{figure*}

\begin{figure*}[htbp]
    \centering
    \includegraphics[width=\linewidth]{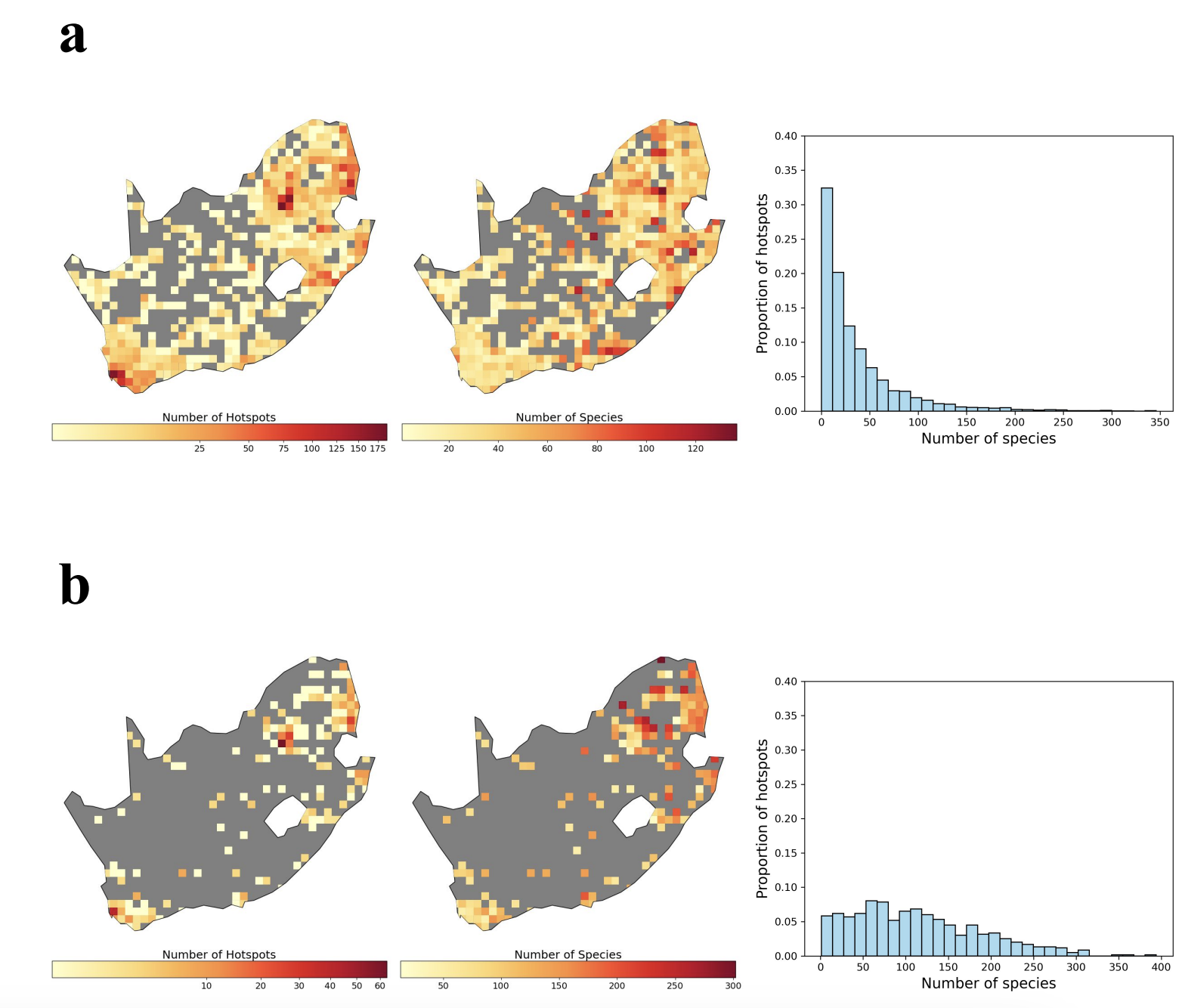}
    \caption{From left to right, for a) Training and b) Test : Geographical distribution of the total number of hotspots (left) and non-zero encounter rates (middle) associated with our South-Africa subdataset, and distribution of the number of species encountered per hotspot (right).}
    \label{figsuppza}
\end{figure*}

\begin{figure*}[htbp]
    \centering
    \includegraphics[width=\linewidth]{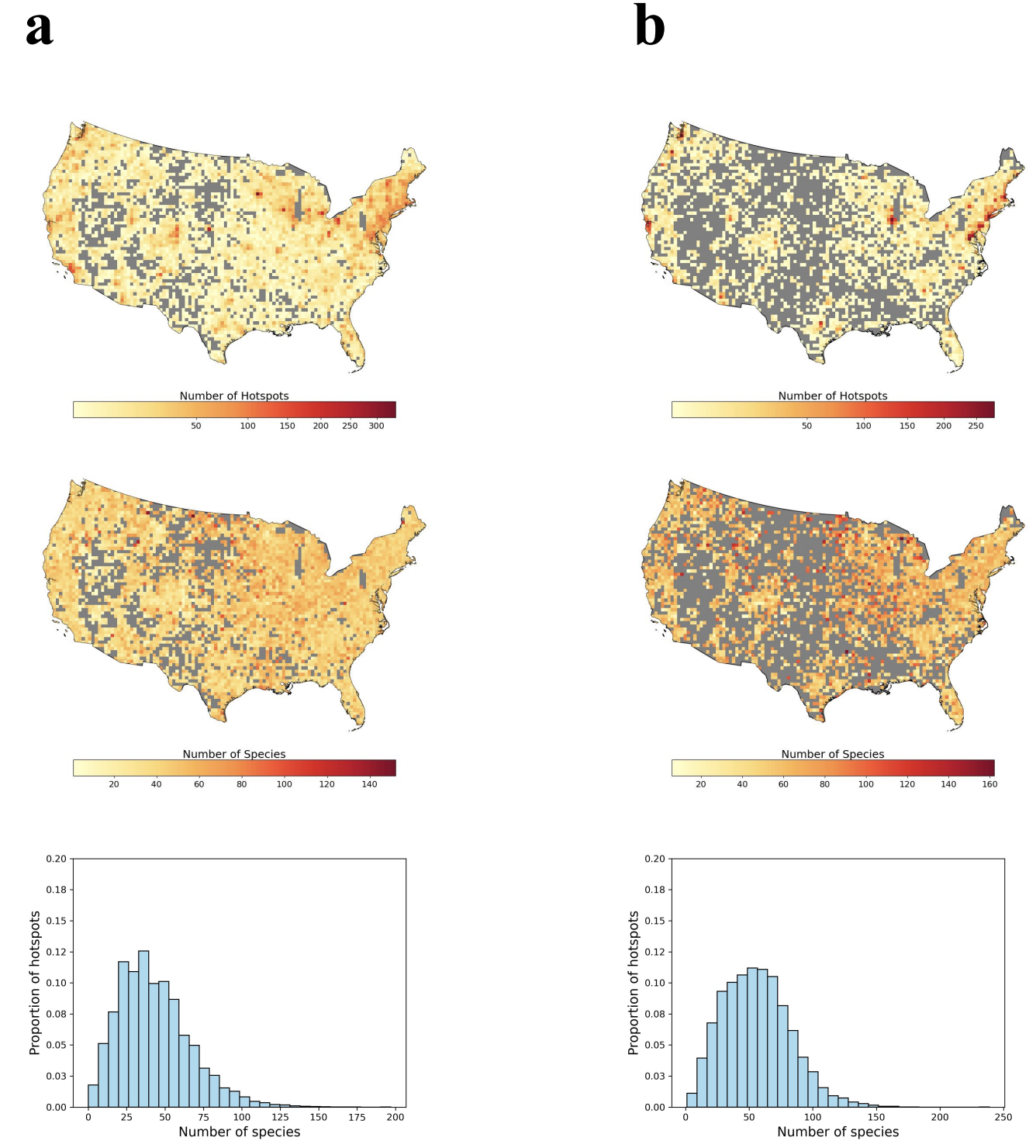}
    \caption{From top to bottom, for a) Training and b) Test : Geographical distribution of the total number of hotspots (top) and non-zero encounter rates (middle) associated with our US-Sumer subdataset, and distribution of the number of species encountered per hotspot (bottom).}
    \label{figsuppsummer}
\end{figure*}

\begin{figure*}[h!]
    \centering
\includegraphics[width=0.99\linewidth]{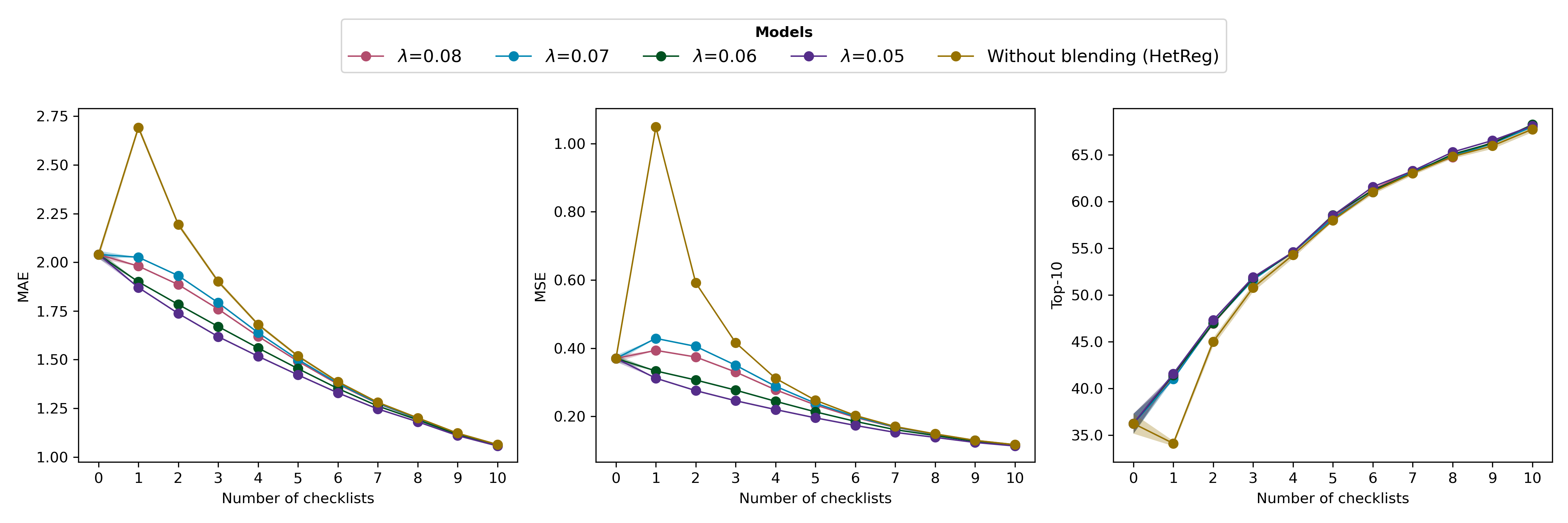}
\caption{Iterative improvements of MAE, MSE and Top-10 metrics on the Kenya sub-dataset with increasing number of BATIS updates for our best performing uncertainty-aware model (HetReg), with and without progressive blending of prior and posterior predictions. Results are shown for various  values for hyperparameter $\lambda$ (see equation \ref{blendingequation}).}
\label{fig:blendingappendix}
\end{figure*}

\section{Limitations}
\label{appendix:limitations}
As shown in Figure \ref{figresiterative}, the initial predictions of an uncertainty-aware SDM can be overwritten in a first Bayesian update, when the estimated prior variance is large. This can lead to an initial drop in performance before enough updates are performed to boost performance, even though performance would ideally increase \textit{monotonically} with number of updates. A simple approach to mitigate such behavior is to consider a weighted average of predictions from the prior and the updated SDM, progressively giving more importance to the updates with more observations. For example, we can consider blended predictions 
\begin{equation}
    p_{blend} = (1-w_{t}) \cdot p_{prior} + w_{t} \cdot p_{posterior}
\label{blendingequation}
\end{equation}
where $w_{t} \in (0, 1]$ is an appropriate discounting factor. A practical choice for $w_{t}$ is $w_{t} = 1 - \exp^{-\lambda t}$, where $t$ corresponds to the number of BATIS posterior updates and $\lambda \in (0, 1]$ is a tunable hyperparameter. Figure \ref{fig:blendingappendix} illustrates how this strategy can progressively smooth performance curves for our best performing uncertainty-aware model (HetReg) on the Kenya sub-dataset.

\section{Methods}
\subsection{Uncertainty-agnostic methods}
\label{appendix:methods-sdm}

We give additional details below on the implementation of our uncertainty-agnostic methods. A summary of the hyperparameter values we used for each of these approaches can be found in Table \ref{tab:agnostichyperparam}. 

\paragraph{Mean Encounter Rate.} As described in the main paper, we simply computed the average of encounter rates over the training set for each species. 

\paragraph{Random Forest.} A Random Forest Regressor is an ensemble learning method that builds multiple decision trees during training and averages their predictions to improve accuracy and reduce overfitting. Each tree is trained on a random subset of the data, in order to diversity among the models. Our Random Forest baseline uses 100 trees, and clips the output between $[0, 1]$ to predict the encounter rate. To implement this baseline, we simply used the \texttt{RandomForestRegressor} class from the the scikit-learn library. 

\paragraph{Multi-Layer Perceptron.} A Multi-Layer Perceptron (MLP) is a simple feedforward neural network consisting of an input layer, one or more hidden layers, and an output layer \cite{deeplearningbook}. We used a single hidden layer of dimension 64 and sigmoid activation layers. The MLP was trained with the binary cross-entropy loss, i.e. 
\begin{equation}
    \mathcal{L}_{CE} := \frac{1}{N} \sum_{i=1}^{N} - \mathbf{y}_{i} \log ( \mathbf{\hat{y}}_{i}) - (1 - \mathbf{y}_{i} ) \log (1 - \hat{\mathbf{y}}_{i} ) 
\label{eq:bceloss}
\end{equation}
where $N$ corresponds to the number of hotspots in a given batch, $\mathbf{y}_{i}$ corresponds the ground encounter rate vector of hotspot $h_{i} \in \mathcal{H}$, and $\hat{\mathbf{y}}_{i}$ corresponds to the predicted encounter rate vector. 
\paragraph{Resnet-18.} We used a conventional Resnet-18 architecture \cite{resnetcvpr}, initialized with weights pre-trained on ImageNet. The model was trained with the binary cross-entropy loss (equation \ref{eq:bceloss}), and implemented in Pytorch with the help of the \texttt{torchvision.models.resnet18} class. 

\begin{table}[h!]
\centering
{\fontsize{9}{\baselineskip}\selectfont
\centering
\begin{tabular}{|lll|}
\Xhline{1.2pt} 
\multicolumn{1}{|c}{\textbf{Parameter}} & \multicolumn{1}{c}{\textbf{Description}}                  & \multicolumn{1}{c|}{\textbf{Assigned value}} \\ \Xhline{1.2pt} 
\multicolumn{3}{|c|}{\cellcolor[HTML]{C0C0C0}Random Forest}                                                                                             \\ \Xhline{1.2pt} 
\multicolumn{1}{|l|}{\texttt{n\_tree}}                & \multicolumn{1}{l|}{Number of trees} & 100                                          \\ \Xhline{1.2pt} 
\multicolumn{3}{|c|}{\cellcolor[HTML]{C0C0C0}Multi-Layer Perceptron}                                                                                    \\ \Xhline{1.2pt} 
\multicolumn{1}{|l|}{\texttt{n\_layers}}              & \multicolumn{1}{l|}{Number of hidden layers}              & 1                                            \\
\multicolumn{1}{|l|}{\texttt{layer\_size}}            & \multicolumn{1}{l|}{Size of hidden layer(s)}              & 64                                           \\ \Xhline{1.2pt} 
\end{tabular}
}
\caption{Summary of hyperparameter values used to train our uncertainty-agnostic methods}
\label{tab:agnostichyperparam}
\end{table}

\subsection{Uncertainty-aware methods}
\label{appendix:methods-uncertainty}

We give additional details below on the implementation of our Monte-Carlo Dropout, Mean-Variance Network and Heteroscedastic Regression Neural Network. Sufficient details are provided in the main paper for the other approaches. A summary of the hyperparameter values we used for each of our uncertainty-aware methods can be found in Table \ref{tab:hyperparamtable}. 

\paragraph{Monte-Carlo Dropout.} Monte-Carlo Dropout (MCD) randomly deactivates a proportion $\rho$ of neurons at training and test time, and uses $M$ forward passes to compute the mean and variance for each hotspot. Dropout layers need to be careful placed, as they can quickly degrade accuracy for large models, which will dramatically increase the computation time required to achieve a satisfying performance \cite{manual2}. We did the same as \cite{hetregpaper} and added only a single dropout layer before the last layer of the network.

\paragraph{Mean-Variance Network.} A Mean-Variance Network (MVN) maps each location to two outputs: a predicted mean encounter rate vector and a predicted variance vector. It does so by using the \textbf{Gaussian negative log-likelihood function}, i.e. 
\begin{equation}
   \mathcal{L}_{GL} := \frac{1}{N} \sum_{i=1}^{N} \frac{1}{2} \log \mathbf{\hat{\mathbf{\sigma}}}_{i}^{2} + \frac{(\mathbf{y}_{i}-\mathbf{\hat{y}}_{i})^{2}}{2\hat{\mathbf{\sigma}}_{i}^{2}}
\end{equation}

where $N$ corresponds to the number of hotspots in a given batch, $\mathbf{y}_{i}$ corresponds to the ground truth encounter rate vector of hotspot $h_{i} \in \mathcal{H}$, and $\mathbf{\hat{y}}_{i}^{2}$ and $\mathbf{\hat{\sigma}}_{i}^{2}$ respectively corresponds to the predicted encounter rate and variance vectors. 

In order to optimize performance, we followed recommendations formulated in \cite{mvapaper1} and added a separate regularization of the mean and the variance
estimate. Our loss function therefore became 
{\fontsize{9}{\baselineskip}\selectfont
\begin{align}
   \mathcal{L}_{GLT} := \frac{1}{N} \sum_{i=1}^{N} \left [ \frac{1}{2} \log \hat{\sigma}_{i}^{2} + \frac{(\mathbf{y}_{i}-\mathbf{\hat{y}}_{i})^{2}}{2\hat{\sigma}_{i}^{2}} \right ] \notag \\ + \lambda_{\mu} \frac{1}{N} \sum \mathbf{\hat{y}}_{i}^{2} + \lambda_{\sigma} \frac{1}{N} \sum \log (\hat{\sigma}_{i}^{2})^{2}
\end{align}
}
where $\lambda_{\mu}$ and $\lambda_{\sigma}$ are tunable hyperparameters. The $\lambda_{\mu}$ parameter penalizes very large predictions, and $\lambda_{\sigma}$ prevent the model from being over or under-confident. 

Another choice we made to optimize training is to add a \textbf{warm-up period} at the beginning, to prevent the model from minimizing the loss by maximizing the variance at the start. We do so by simply setting $\hat{\sigma}_{i}^{2} = 1$ for a pre-determined number of epochs at the start. 

\paragraph{Heteroscedastic Regression Neural Network \cite{hetregpaper}.} A Heteroscedastic Regression Neural Network (HetReg) is very similar to an MVN, but adds MC Dropout sampling to simultaneously quantify aleatoric and epistemic uncertainty. $M$ dropout passes are used to compute the mean encounter rate vector, and variance is estimated by adding epistemic uncertainty (variance computed from $M$ predicted encounter rates) to aleatoric uncertainty (mean computed from $M$ predicted variances). 

We applied the same optimizations to HetReg than to MVN, and we implemented dropout the same way as we did for our MCD baseline. 

\begin{table*}[h!]
{\fontsize{9}{\baselineskip}\selectfont
\renewcommand{\arraystretch}{1.0}
\begin{tabular}{lll|}
\Xhline{1.2pt} 
\multicolumn{1}{|c}{\textbf{Hyperparameter}} & \multicolumn{1}{c}{\textbf{Description}}                                                                        & \multicolumn{1}{c|}{\textbf{Assigned value}} \\ \Xhline{1.2pt} 
\multicolumn{3}{|c|}{\cellcolor[HTML]{C0C0C0}Resnet-18 + Fixed Variance}                                                                                                                                                    \\ \Xhline{1.2pt} 
\multicolumn{1}{|l|}{$\tau$}                 & \multicolumn{1}{l|}{Scaling Parameter for the Variance}                                                         & 1                                            \\ \Xhline{1.2pt} 
\multicolumn{3}{|c|}{\cellcolor[HTML]{C0C0C0}Resnet-18+Historical Variance}                                                                                                                                                    \\ \Xhline{1.2pt} 
\multicolumn{1}{|l|}{\texttt{edge\_case}}           & \multicolumn{1}{l|}{\begin{tabular}[c]{@{}l@{}}Value to assign to variance when historical variance is too high\end{tabular}} & $\mu \cdot (1 - \mu)$                        \\ \Xhline{1.2pt} 
\multicolumn{3}{|c|}{\cellcolor[HTML]{C0C0C0}Resnet-18 + Mean-Variance Network}                                                                                                                                                   \\ \Xhline{1.2pt} 
\multicolumn{1}{|l|}{$\lambda_{\mu}$}        & \multicolumn{1}{l|}{Regularization parameter for predicted mean}                                                & 0.1                                          \\
\multicolumn{1}{|l|}{$\lambda_{\sigma}$}     & \multicolumn{1}{l|}{Regularization parameter for predicted variance}                                            & 0.1                                          \\
\multicolumn{1}{|l|}{\texttt{warmup}}                 & \multicolumn{1}{l|}{Number of epochs to warmup model}                                                           & 5                                            \\ \Xhline{1.2pt} 
\multicolumn{3}{c|}{\cellcolor[HTML]{C0C0C0}Resnet-18 + Dropout}                                                                                                                                                    \\ \Xhline{1.2pt} 
\multicolumn{1}{|l|}{$\rho$}                 & \multicolumn{1}{l|}{Dropout rate}                                                                               & 0.2                                          \\
\multicolumn{1}{|l|}{$M$}                      & \multicolumn{1}{l|}{Number of dropout passes to compute predicted mean and variance}                            & 30                                           \\ \Xhline{1.2pt} 
\multicolumn{3}{|c|}{\cellcolor[HTML]{C0C0C0}Resnet-18 + HetReg}                                                                                                                                                \\ \Xhline{1.2pt} 
\multicolumn{1}{|l|}{$\lambda_{\mu}$}        & \multicolumn{1}{l|}{Regularization parameter for mean}                                                          & 0.1                                          \\
\multicolumn{1}{|l|}{$\lambda_{\sigma}$}     & \multicolumn{1}{l|}{Regularization parameter for variance}                                                      & 0.1                                          \\
\multicolumn{1}{|l|}{\texttt{warmup}}                 & \multicolumn{1}{l|}{Number of epochs to warmup model}                                                           & 5                                            \\
\multicolumn{1}{|l|}{$\rho$}                 & \multicolumn{1}{l|}{Dropout rate}                                                                               & 0.2                                          \\
\multicolumn{1}{|l|}{$M$}                      & \multicolumn{1}{l|}{Number of dropout passes to compute predicted mean and variance}                            & 30                                           \\ \Xhline{1.2pt} 
\multicolumn{3}{|c|}{\cellcolor[HTML]{C0C0C0}Resnet-18 + Deep Ensembles}                                                                                                                                                    \\ \Xhline{1.2pt} 
\multicolumn{1}{|l|}{$M$}                      & \multicolumn{1}{l|}{Number of independently trained models from which to compute predicted mean and variance}   & 5                                            \\ \Xhline{1.2pt} 
\multicolumn{3}{|c|}{\cellcolor[HTML]{C0C0C0}Resnet-18 + Shallow Ensembles}                                                                                                                                                    \\ \Xhline{1.2pt} 
\multicolumn{1}{|l|}{$M$}                      & \multicolumn{1}{l|}{Number of heads}                                                                            & 5                                            \\ \Xhline{1.2pt} 
\end{tabular}
\caption{Summary of hyperparameter values used to train our uncertainty-aware methods}
\label{tab:hyperparamtable}
}
\end{table*}

\section{Experiments}
\label{appendix:experiments}
\subsection{Training protocol.}\label{appendix:trainingprotocol}
All deep learning models were trained using the Adam \cite{kingma2014adam} optimizer, with a batch size of 128 and for a maximum of 50 epochs. We trained each model with three different seeds, and we used 30 dropout pass at test time to compute the mean and variance of predictions for the approaches requiring a dropout layer to quantify uncertainty \cite{pmlr-v48-gal16}. Random noise, blur and vertical/horizontal flipping were used for data augmentation, and we normalized inputs (satellite imagery and/or bioclimatic variables) using training set statistics. The optimizer hyperparameter values we used in each of our experiments are listed in Table \ref{tab:learningprotocolhyperparam}. In addition, a summary of the hyperparameter values we used for each of our uncertainty-aware methods can be found in Table \ref{tab:hyperparamtable}. 

 We report results generated with the same learning rate ($0.0001$) for all Resnet18-based models because it was demonstrated to be the one that gave the best results in the related SatBird task \cite{NEURIPS2023_ef7653bb}. For our Monte-Carlo Dropout approach, we took inspiration from the experiments in \citet{pmlr-v48-gal16} and considered $\rho = 0.2$ as the dropout rate for MCD and HetReg. For the \texttt{warmup}, $\lambda_{\mu}$ and $\lambda_{\sigma}$ hyperparameters of our Mean-Variance Network and HetReg approaches, we  considered $\text{\texttt{warmup}} = \left[2, 5, 10 \right]$, $\lambda_{\mu} = \left[0.1, 0.001, 0.0001 \right]$, $\lambda_{\sigma} = \left[0.1, 0.001, 0.0001 \right]$, and found that using $\text{\texttt{warmup}} = 5$, $\lambda_{\mu} = 0.1$ and $\lambda_{\sigma} = 0.1$ led to the best results. 
 
\paragraph{Random Seeds.} For each individual experiment, we used a random number generator to generate on seed number, which was then used to set \texttt{numpy.random.seed}, \texttt{random.seed} and \texttt{torch.manual\_seed}. The generated seed can then be re-used for reproducibility of a single experiment if needed. For each model in our benchmark, we showed results averaged from 3 different random seeds (i.e., we trained each model 3 different times). 

\begin{table*}[h!]
\centering
{\fontsize{9}{\baselineskip}\selectfont
\renewcommand{\arraystretch}{1.0}
\begin{tabular}{|lll|}
\Xhline{1.2pt}
\multicolumn{1}{|c}{\textbf{Hyperparameter}} & \multicolumn{1}{c}{\textbf{Description}}                                                                                                                       & \multicolumn{1}{c|}{\textbf{Assigned value}} \\ \Xhline{1.2pt}
\multicolumn{3}{|c|}{\cellcolor[HTML]{C0C0C0}Multi-Layer Perceptron}                                                                                                                                                                                         \\ \Xhline{1.2pt}
\multicolumn{1}{|l|}{\texttt{learning\_rate}}      & \multicolumn{1}{l|}{Learning Rate (for Adam)}                                                                                                                  & 0.001                                        \\
\multicolumn{1}{|l|}{\texttt{batch\_size}}            & \multicolumn{1}{l|}{Batch size}                                                                                                                                & 128                                          \\
\multicolumn{1}{|l|}{\texttt{max\_epochs}}         & \multicolumn{1}{l|}{Maximum number of epochs to train the model}                                                                                               & 50                                           \\
\multicolumn{1}{|l|}{\texttt{factor}}                 & \multicolumn{1}{l|}{\begin{tabular}[c]{@{}l@{}}Reduction factor for the learning rate if validation loss \\ does not improve for patience epochs\end{tabular}} & 0.5                                          \\
\multicolumn{1}{|l|}{\texttt{patience}}               & \multicolumn{1}{l|}{Number of successive epochs to monitor change in validation loss}                                                                          & 20                                           \\ \Xhline{1.2pt}
\multicolumn{3}{|c|}{\cellcolor[HTML]{C0C0C0}Resnet18-based Models}                                                                                                                                                                                          \\ \Xhline{1.2pt}
\multicolumn{1}{|l|}{\texttt{learning\_rate}}         & \multicolumn{1}{l|}{Number of hidden layers (for Adam)}                                                                                                        & 0.0001                                       \\
\multicolumn{1}{|l|}{\texttt{weight\_decay}}          & \multicolumn{1}{l|}{L2 Regularization (for Adam)}                                                                                                              & 0.00001                                      \\
\multicolumn{1}{|l|}{\texttt{batch\_size}}            & \multicolumn{1}{l|}{Maximum number of epochs to train the model}                                                                                               & 128                                          \\
\multicolumn{1}{|l|}{\texttt{max\_epochs}}            & \multicolumn{1}{l|}{Maximum number of epochs to train the model}                                                                                               & 50                                           \\
\multicolumn{1}{|l|}{\texttt{factor}}                 & \multicolumn{1}{l|}{\begin{tabular}[c]{@{}l@{}}Reduction factor for the learning rate if validation loss \\ does not improve for patience epochs\end{tabular}} & 0.5                                          \\
\multicolumn{1}{|l|}{\texttt{patience}}               & \multicolumn{1}{l|}{Number of successive epochs to monitor change in validation loss}                                                                          & 20                                           \\ \Xhline{1.2pt}
\end{tabular}
}
\caption{Summary of hyperparameter values used to parametrize our model training protocol}
\label{tab:learningprotocolhyperparam}
\end{table*}

\subsection{Metrics}\label{appendix:metrics}

Let $N$ be the number of species, $\mathcal{H}$ be a set of hotspots, $y_{i}$ be the ground truth encounter rate vector for species $i$ at hotspot $h \in \mathcal{H}$, $\hat{y}_{i}$ be the predicted encounter rate for species $i$ at hotspot $h \in \mathcal{H}$. Our metrics are defined as follows : 

\paragraph{Mean Absolute Error (MAE).} The MAE for a single hotspot $h \in \mathcal{H}$ is defined as 
\begin{equation}
    \text{MAE} := \frac{1}{N} \sum_{i=1}^{N} | y_{i} - \hat{y}_{i} |
\end{equation}

\paragraph{Mean Squared Error (MSE).} The MSE for a single hotspot $h \in \mathcal{H}$ is defined as 
\begin{equation}
    \text{MSE} := \frac{1}{N} \sum_{i=1}^{N} (y_{i} - \hat{y}_{i})^{2}
\end{equation}

\paragraph{Adaptive Top-$\mathbf{k}$.} Our Top-$k$ metric measures how well the model's top-$k$ predicted species (those it expects to be the most frequently encountered) align with the ground truth top-$k$ species encountered in the field. This metric can be formally defined as 
\begin{equation}
    \text{Top-k} := \begin{cases}
        \frac{| \mathcal{I}_{\hat{y}} \bigcap \mathcal{I}_{y} |}{k}\cdot100 & \text{if $k > 0$} \\
        0 & \text{otherwise}
    \end{cases}
\end{equation}
where $k = | {i : y_{i} \neq 0} |$ denotes the number of non-zero ground truth encounter rates at hotspot $h \in \mathcal{H}$, $\mathcal{I}_{\hat{y}}$ denotes the sets of indices of the top-$k$ largest values in $\mathbf{\hat{y}}$, and $\mathcal{I}_{y}$ denotes the sets of indices of the top-$k$ largest values in $\mathbf{y}$. 

A top-$k$ score of 100\% means the model perfectly identified all of the $k$ most commonly encountered species at hotspot $h \in \mathcal{H}$, and a score of 0 indicates that none of the top-$k$ predicted species matched the ground truth top-$k$ species.

\paragraph{Top-10 and Top-30.} Adaptive Top-10 and Top-30 are defined similarly as top-$k$, except that $k$ is respectively replaced by 10 or 30. 
\subsection{Python Environment}
\label{appendix:pythonenv}
Our code was developed in Python 3.10. The required packages can be installed using the \texttt{requirements.txt} file provided in our Github repository.

\begin{table*}[!ht]
\centering
{\fontsize{9}{\baselineskip}\selectfont
\renewcommand{\arraystretch}{1.0}
\begin{tabular}{|lcccccccccccc|}
\hline
\multicolumn{1}{|l|}{}                   & \multicolumn{1}{l}{} & \multicolumn{10}{c}{\textbf{Subset}}                                                                                                                                & \multicolumn{1}{l|}{} \\ \cline{3-12}
\multicolumn{1}{|c|}{\textbf{Resources}} & \multicolumn{1}{l}{} & \textbf{Kenya} &                       &  & \textbf{South-Africa} &                       &  & \textbf{USA-Winter} &                       &  & \textbf{USA-Summer} & \multicolumn{1}{l|}{} \\ \hline
\multicolumn{13}{|c|}{\cellcolor[HTML]{C0C0C0}\begin{tabular}[c]{@{}c@{}}Resnet-18, Resnet+HetReg, Resnet+MVN, Resnet+MCD, Resnet+SE, Resnet+DE\end{tabular}}                                                                      \\ \hline
\multicolumn{1}{|l|}{\texttt{time}}               &                      & 10:00:00       & \multicolumn{1}{c|}{} &  & 12:00:00              & \multicolumn{1}{c|}{} &  & 23:59:00            & \multicolumn{1}{c|}{} &  & 23:59:00            &                       \\
\multicolumn{1}{|l|}{\texttt{node}}               &                      & 1              & \multicolumn{1}{c|}{} &  & 1                     & \multicolumn{1}{c|}{} &  & 1                   & \multicolumn{1}{c|}{} &  & 1                   &                       \\
\multicolumn{1}{|l|}{\texttt{mem}}                &                      & 24G            & \multicolumn{1}{c|}{} &  & 24G                   & \multicolumn{1}{c|}{} &  & 24G                 & \multicolumn{1}{c|}{} &  & 24G                 &                       \\
\multicolumn{1}{|l|}{\texttt{cpus-per-task}}      &                      & 1              & \multicolumn{1}{c|}{} &  & 1                     & \multicolumn{1}{c|}{} &  & 1                   & \multicolumn{1}{c|}{} &  & 1                   &                       \\
\multicolumn{1}{|l|}{\texttt{gpus-per-node}}      &                      & 1              & \multicolumn{1}{c|}{} &  & 1                     & \multicolumn{1}{c|}{} &  & 1                   & \multicolumn{1}{c|}{} &  & 1                   &                       \\ \hline
\end{tabular}
}
\caption{Summary of computing resources required by our job scripts for the four subsets of our dataset. Our Mean Encounter Rate, Random Forest and Multi-Layer Perceptron baselines were not trained on SLURM-based clusters.}
\label{tab:computingresources}
\end{table*}

\subsection{Execution Environment}
\label{appendix:executionenv}
We validated that all our code can be executed on :
\begin{itemize}
    \item SLURM-based clusters
    \item macOS Sonoma 14.1 (with Apple M Series chips)
    \item  Windows 11 (with NVIDIA GeForce RTX 4070)
\end{itemize}

\paragraph{Model Training (Resnet-18).} Resnet-18 models were trained independently on a computer cluster equipped with NVIDIA V100 GPUs (16 GB). The cluster we used rely on the open-source \texttt{SLURM} job scheduler, which is used by many of the world's supercomputers. Our code environment should therefore work in other similar computer clusters. A  summary of the computing resources our job scripts are requiring can be found in Table \ref{tab:computingresources}. 

\paragraph{Model Training (Baselines).} We trained our Mean Encounter Rate, Random Forest and Multi-Layer Perceptron baselines on a MacBook Pro equipped with an Apple M3 Pro Chip (18 GB), running on macOS Sonoma 14.1. 

\paragraph{Bayesian Updating Framework.} We performed the Bayesian Updating Framework experiments on a MacBook Pro equipped with an Apple M3 Pro Chip (18 GB), running on macOS Sonoma 14.1.  

\section{Results}
\label{app:results} 

\subsection{Additional Results for Kenya}
\paragraph{Iterative Improvements by Region}. Figure \ref{fig:kenyagraphsappendix} shows the evolution of the performance of uncertainty-aware approaches with increasing number of checklist updates, on the Kenya subdataset. We can observe the same trends as Figure \ref{figresiterative} of the main paper. 

\subsection{Results for USA-Summer}

\paragraph{Overall.} Table \ref{resusasummer} shows results for the USA-Winter subdataset. 

\paragraph{Iterative Improvements by Region}. Figure \ref{fig:usasummergraphsappendix} shows the evolution of the performance of uncertainty-aware approaches with increasing number of checklist updates, on the USA-Summer subdataset. We can observe the same trends as Figure \ref{figresiterative} of the main paper. 

\begin{figure*}[h!]
    \centering
\includegraphics[width=0.99\linewidth]{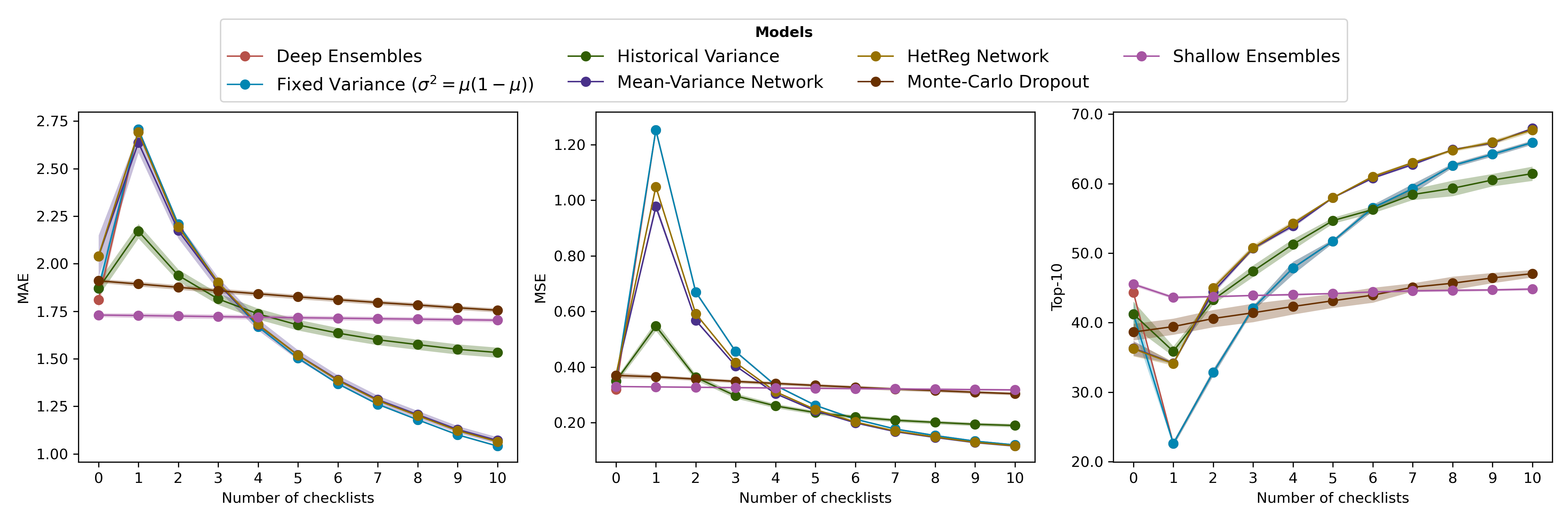}
\caption{Iterative improvements for the different uncertainty estimation approaches with increasing
number of checklist updates for the MAE, MSE and Top-10 metrics on the Kenya Region test
set. We report the mean on three seeds and standard deviations for each model.}
\label{fig:kenyagraphsappendix}
\end{figure*}

\begin{figure*}[h!]
    \centering
\includegraphics[width=0.99\linewidth]{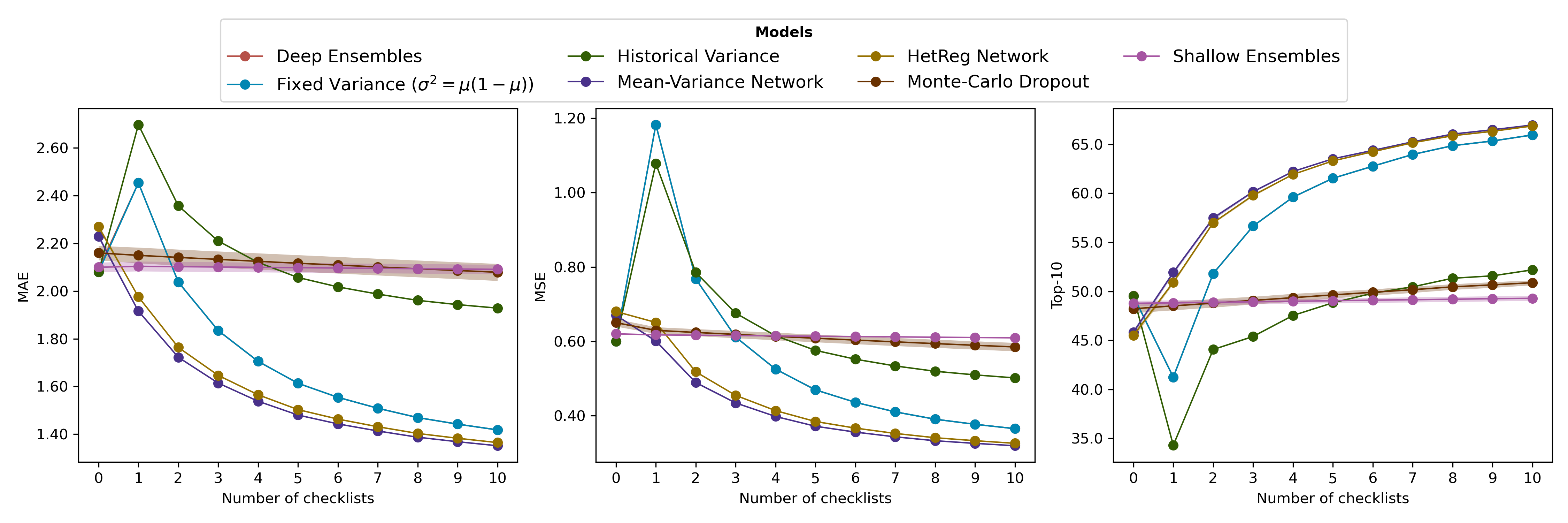}
    \caption{Iterative improvements for the different uncertainty estimation approaches with increasing
number of checklist updates for the MAE, MSE and Top-10 metrics on the USA-Summer Region test
set. We report the mean on three seeds and standard deviations for each model.}
\label{fig:usasummergraphsappendix}
\end{figure*} 

\begin{table*}[h!]
    \centering
\begin{minipage}{0.95\linewidth}
\centering
\begin{subtable}[t]{\linewidth}
\centering
{\fontsize{9}{\baselineskip}\selectfont
\setlength{\tabcolsep}{1mm}
\begin{tabular}{|llllccccccccccl|}
        \toprule
 & \multicolumn{1}{c}{}               & \multicolumn{1}{l|}{} &  & \multicolumn{10}{c}{\textbf{Metrics}}                                                                                                    &  \\ \cline{4-15}
 & \multicolumn{1}{c}{\textbf{Model}} & \multicolumn{1}{l|}{}& &&&&&&&&&&&  \\ 
 &                                    & \multicolumn{1}{l|}{} &  & \multicolumn{2}{c}{MAE[1e-2]} & \multicolumn{2}{c}{MSE[1e-2]} & \multicolumn{2}{c}{Top-10} & \multicolumn{2}{c}{Top-30} & \multicolumn{2}{c}{Top-k} &  \\ \hline\hline

 \multicolumn{3}{|c}{\cellcolor[HTML]{D3D3D3}}&
\multicolumn{12}{c|}{\cellcolor[HTML]{D3D3D3}\textbf{USA-Summer}}                                                                                                                                                \\ \hline
  & Mean Encounter Rate               & \multicolumn{1}{l|}{} &  &           \multicolumn{2}{c|}{\makecell{\colorone{3.08}\\\textcolor{gray}{$\pm$0.00}}} & \multicolumn{2}{c|}{\makecell{\colorone{0.91}\\\textcolor{gray}{$\pm$0.00}}}  &  \multicolumn{2}{c|}{\makecell{\colorone{29.69}\\\textcolor{gray}{$\pm$0.00}}} &   \multicolumn{2}{c|}{\makecell{\colorone{45.47}\\\textcolor{gray}{$\pm$0.00}}}                           &  \multicolumn{2}{c}{\makecell{\colorone{51.97}\\\textcolor{gray}{$\pm$0.00}}}                                      &  \\
  & MLP          & \multicolumn{1}{l|}{} &  &      \multicolumn{2}{c|}{\makecell{\colorone{2.29}\\\textcolor{gray}{$\pm$0.01}}} & \multicolumn{2}{c|}{\makecell{\colorone{0.66}\\\textcolor{gray}{$\pm$0.00}}}  &  \multicolumn{2}{c|}{\makecell{\colorone{44.18}\\\textcolor{gray}{$\pm$0.12}}} &   \multicolumn{2}{c|}{\makecell{\colorone{62.27}\\\textcolor{gray}{$\pm$0.03}}}                           &  \multicolumn{2}{c}{\makecell{\colorone{69.04}\\\textcolor{gray}{$\pm$0.01}}}                           &  \\
  & Random Forest                      & \multicolumn{1}{l|}{} &  &     \multicolumn{2}{c|}{\makecell{\colorone{1.96}\\\textcolor{gray}{$\pm$0.00}}} & \multicolumn{2}{c|}{\makecell{\colorone{0.59}\\\textcolor{gray}{$\pm$0.00}}}  &  \multicolumn{2}{c|}{\makecell{\colorone{49.38}\\\textcolor{gray}{$\pm$0.04}}} &   \multicolumn{2}{c|}{\makecell{\colorone{67.07}\\\textcolor{gray}{$\pm$0.02}}}                           &  \multicolumn{2}{c}{\makecell{\colorone{72.44}\\\textcolor{gray}{$\pm$0.01}}}                           &  \\
   & ResNet-18                      & \multicolumn{1}{l|}{} &  &     \multicolumn{2}{c|}{\makecell{\colorone{2.08}\\\textcolor{gray}{$\pm$0.00}}} & \multicolumn{2}{c|}{\makecell{\colorone{0.60}\\\textcolor{gray}{$\pm$0.00}}}  &  \multicolumn{2}{c|}{\makecell{\colorone{49.52}\\\textcolor{gray}{$\pm$0.05}}} &   \multicolumn{2}{c|}{\makecell{\colorone{66.28}\\\textcolor{gray}{$\pm$0.01}}}                           &  \multicolumn{2}{c}{\makecell{\colorone{78.45}\\\textcolor{gray}{$\pm$0.01}}}                           &  \\\hline\hline
 & ResNet-18+FV                      & \multicolumn{1}{l|}{} &  &     \makecell{\colorone{2.08}\\\textcolor{gray}{$\pm$0.00}} & \multicolumn{1}{c|}{\makecell{\colortwo{1.61}\\\textcolor{gray}{$\pm$0.00}}}  &  \makecell{\colorone{0.60}\\\textcolor{gray}{$\pm$0.00}} &   \multicolumn{1}{c|}{\makecell{\colortwo{0.47}\\\textcolor{gray}{$\pm$0.00}}}                           &  \makecell{\colorone{49.52}\\\textcolor{gray}{$\pm$0.05}}                           &  
  \multicolumn{1}{c|}{\makecell{\colortwo{61.53}\\\textcolor{gray}{$\pm$0.01}}} & \makecell{\colorone{66.28}\\\textcolor{gray}{$\pm$0.01}}  &  \multicolumn{1}{c|}{\makecell{\colortwo{75.69}\\\textcolor{gray}{$\pm$0.01}}} &   \makecell{\colorone{78.45}\\\textcolor{gray}{$\pm$0.01}}                           &  \multicolumn{1}{c}{\makecell{\colortwo{80.49}\\\textcolor{gray}{$\pm$2=00.00}}}                           &\\
 & ResNet-18+HV                        & \multicolumn{1}{l|}{} &  &        \makecell{\colorone{2.08}\\\textcolor{gray}{$\pm$0.00}} & \multicolumn{1}{c|}{\makecell{\colortwo{2.06}\\\textcolor{gray}{$\pm$0.00}}}  &\makecell{\colorone{0.60}\\\textcolor{gray}{$\pm$0.00}}  &  \multicolumn{1}{c|}{\makecell{\colortwo{0.57}\\\textcolor{gray}{$\pm$0.00}}}  &\makecell{\colorone{49.52}\\\textcolor{gray}{$\pm$0.04}} &\multicolumn{1}{c|}{\makecell{\colortwo{48.83}\\\textcolor{gray}{$\pm$0.04}}}  &   \makecell{\colorone{66.28}\\\textcolor{gray}{$\pm$0.01}}                           & \multicolumn{1}{c|}{\makecell{\colortwo{69.79}\\\textcolor{gray}{$\pm$0.02}}}  & \makecell{\colorone{78.45}\\\textcolor{gray}{$\pm$0.01}}                           &  \multicolumn{1}{c}{\makecell{\colortwo{75.56}\\\textcolor{gray}{$\pm$0.02}}}                           &\\
 & ResNet-18+DE                        & \multicolumn{1}{l|}{} &  &     \makecell{\colorone{2.09}\\\textcolor{gray}{$\pm$0.00}} & \multicolumn{1}{c|}{\makecell{\colortwo{2.05}\\\textcolor{gray}{$\pm$0.06}}}  &  \makecell{\colorone{0.60}\\\textcolor{gray}{$\pm$0.01}} &   \multicolumn{1}{c|}{\makecell{\colortwo{0.57}\\\textcolor{gray}{$\pm$0.00}}}                           &  \makecell{\colorone{49.54}\\\textcolor{gray}{$\pm$0.02}}                           &  
  \multicolumn{1}{c|}{\makecell{\colortwo{50.81}\\\textcolor{gray}{$\pm$0.01}}} & \makecell{\colorone{66.34}\\\textcolor{gray}{$\pm$0.01}}  &  \multicolumn{1}{c|}{\makecell{\colortwo{67.37}\\\textcolor{gray}{$\pm$0.01}}} &   \makecell{\colorone{78.47}\\\textcolor{gray}{$\pm$0.01}}                           &  \multicolumn{1}{c}{\makecell{\colortwo{78.81}\\\textcolor{gray}{$\pm$0.00}}}                           &\\
 & ResNet-18+SE                        & \multicolumn{1}{l|}{} &  &     \makecell{\colorone{2.10}\\\textcolor{gray}{$\pm$0.02}} & \multicolumn{1}{c|}{\makecell{\colortwo{2.09}\\\textcolor{gray}{$\pm$0.02}}}  &  \makecell{\colorone{0.62}\\\textcolor{gray}{$\pm$0.00}} &   \multicolumn{1}{c|}{\makecell{\colortwo{0.61}\\\textcolor{gray}{$\pm$0.00}}}                           &  \makecell{\colorone{48.78}\\\textcolor{gray}{$\pm$0.27}}                           &  
  \multicolumn{1}{c|}{\makecell{\colortwo{49.04}\\\textcolor{gray}{$\pm$0.25}}} & \makecell{\colorone{65.62}\\\textcolor{gray}{$\pm$0.21}}  &  \multicolumn{1}{c|}{\makecell{\colortwo{65.81}\\\textcolor{gray}{$\pm$0.19}}} &   \makecell{\colorone{78.04}\\\textcolor{gray}{$\pm$0.15}}                           &  \multicolumn{1}{c}{\makecell{\colortwo{78.08}\\\textcolor{gray}{$\pm$0.14}}}                           &\\ 
  & ResNet-18+MCD                        & \multicolumn{1}{l|}{} &  &     \makecell{\colorone{2.16}\\\textcolor{gray}{$\pm$0.03}} & \multicolumn{1}{c|}{\makecell{\colortwo{2.11}\\\textcolor{gray}{$\pm$0.03}}}  &  \makecell{\colorone{0.65}\\\textcolor{gray}{$\pm$0.01}} &   \multicolumn{1}{c|}{\makecell{\colortwo{0.60}\\\textcolor{gray}{$\pm$0.01}}}                           &  \makecell{\colorone{48.22}\\\textcolor{gray}{$\pm$0.43}}                           &  
  \multicolumn{1}{c|}{\makecell{\colortwo{49.63}\\\textcolor{gray}{$\pm$0.33}}} & \makecell{\colorone{64.83}\\\textcolor{gray}{$\pm$0.00}}  &  \multicolumn{1}{c|}{\makecell{\colortwo{66.05}\\\textcolor{gray}{$\pm$0.61}}} &   \makecell{\colorone{77.43}\\\textcolor{gray}{$\pm$0.26}}                           &  \multicolumn{1}{c}{\makecell{\colortwo{77.81}\\\textcolor{gray}{$\pm$0.35}}}                           &\\ 
  & ResNet-18+MVN                        & \multicolumn{1}{l|}{} &  &     \makecell{\colorone{2.23}\\\textcolor{gray}{$\pm$0.01}} & \multicolumn{1}{c|}{\makecell{\colortwo{1.48}\\\textcolor{gray}{$\pm$0.00}}}  &  \makecell{\colorone{0.67}\\\textcolor{gray}{$\pm$0.00}} &   \multicolumn{1}{c|}{\makecell{\colortwo{0.37}\\\textcolor{gray}{$\pm$0.00}}}                           &  \makecell{\colorone{45.83}\\\textcolor{gray}{$\pm$0.24}}                           &  
  \multicolumn{1}{c|}{\makecell{\colortwo{63.51}\\\textcolor{gray}{$\pm$0.02}}} & \makecell{\colorone{62.77}\\\textcolor{gray}{$\pm$0.04}}  &  \multicolumn{1}{c|}{\makecell{\colortwo{76.02}\\\textcolor{gray}{$\pm$0.04}}} &   \makecell{\colorone{76.18}\\\textcolor{gray}{$\pm$0.02}}                           &  \multicolumn{1}{c}{\makecell{\colortwo{78.14}\\\textcolor{gray}{$\pm$0.11}}}                           &\\ 
  & ResNet-18+HetReg                    & \multicolumn{1}{l|}{} &  &     \makecell{\colorone{2.27}\\\textcolor{gray}{$\pm$0.01}} & \multicolumn{1}{c|}{\makecell{\colortwo{1.50}\\\textcolor{gray}{$\pm$0.00}}}  &  \makecell{\colorone{0.68}\\\textcolor{gray}{$\pm$0.00}} &   \multicolumn{1}{c|}{\makecell{\colortwo{0.38}\\\textcolor{gray}{$\pm$0.00}}}                           &  \makecell{\colorone{45.49}\\\textcolor{gray}{$\pm$0.29}}                           &  
  \multicolumn{1}{c|}{\makecell{\colortwo{63.22}\\\textcolor{gray}{$\pm$0.12}}} & \makecell{\colorone{62.23}\\\textcolor{gray}{$\pm$0.21}}  &  \multicolumn{1}{c|}{\makecell{\colortwo{75.79}\\\textcolor{gray}{$\pm$0.02}}} &   \makecell{\colorone{75.75}\\\textcolor{gray}{$\pm$0.08}}                           &  \multicolumn{1}{c}{\makecell{\colortwo{77.23}\\\textcolor{gray}{$\pm$0.10}}}                           &\\ 
  \hline 

\end{tabular}
}
\end{subtable}
\end{minipage}
    \caption{Performance of SDM-estimation approaches both \colorone{without Bayesian updates} (left side in split columns) and \colortwo{after updates from five checklists} (right side). Note that the first four methods for each region are not uncertainty-aware and therefore cannot be updated with checklist information. Means are shown $\pm$ standard deviation across runs. (Note that many standard deviations are quite low; those reported as 0.00 are accurate.)}
    \label{resusasummer}
\end{table*}

\subsection{Results for USA-Winter}

\paragraph{Overall.} Table \ref{resusawinter} shows results for the USA-Winter subdataset. 

\paragraph{Iterative Improvements by Region}. Figure \ref{fig:usawintergraphsappendix} shows the evolution of the performance of uncertainty-aware approaches with increasing number of checklist updates, on the Kenya subdataset. We can observe the same trends as Figure \ref{figresiterative} of the main paper.

\begin{table*}[h!]
    \centering
\begin{minipage}{0.95\linewidth}
\centering
\begin{subtable}[t]{\linewidth}
\centering
{\fontsize{9}{\baselineskip}\selectfont
\renewcommand{\arraystretch}{1.0}
\begin{tabular}{|llllccccccccccl|}
        \toprule
 & \multicolumn{1}{c}{}               & \multicolumn{1}{l|}{} &  & \multicolumn{10}{c}{\textbf{Metrics}}                                                                                                    &  \\ \cline{4-15}
 & \multicolumn{1}{c}{\textbf{Model}} & \multicolumn{1}{l|}{}& &&&&&&&&&&&  \\ 
 &                                    & \multicolumn{1}{l|}{} &  & \multicolumn{2}{c}{MAE[1e-2]} & \multicolumn{2}{c}{MSE[1e-2]} & \multicolumn{2}{c}{Top-10} & \multicolumn{2}{c}{Top-30} & \multicolumn{2}{c}{Top-k} &  \\ \hline\hline

 \multicolumn{3}{|c}{\cellcolor[HTML]{D3D3D3}}&
\multicolumn{12}{c|}{\cellcolor[HTML]{D3D3D3}\textbf{USA-Winter}}                                                                                                                                                \\ \hline
  & Mean Encounter Rate               & \multicolumn{1}{l|}{} &  &           \multicolumn{2}{c|}{\makecell{\colorone{2.47}\\\textcolor{gray}{\scriptsize$\pm$0.00}}} & \multicolumn{2}{c|}{\makecell{\colorone{0.71}\\\textcolor{gray}{\scriptsize$\pm$0.00}}}  &  \multicolumn{2}{c|}{\makecell{\colorone{28.75}\\\textcolor{gray}{\scriptsize$\pm$0.00}}} &   \multicolumn{2}{c|}{\makecell{\colorone{50.06}\\\textcolor{gray}{\scriptsize$\pm$0.00}}}                           &  \multicolumn{2}{c}{\makecell{\colorone{55.20}\\\textcolor{gray}{\scriptsize$\pm$0.00}}}                                      &  \\
  & MLP          & \multicolumn{1}{l|}{} &  &      \multicolumn{2}{c|}{\makecell{\colorone{1.96}\\\textcolor{gray}{\scriptsize$\pm$0.01}}} & \multicolumn{2}{c|}{\makecell{\colorone{0.55}\\\textcolor{gray}{\scriptsize$\pm$0.00}}}  &  \multicolumn{2}{c|}{\makecell{\colorone{45.25}\\\textcolor{gray}{\scriptsize$\pm$0.14}}} &   \multicolumn{2}{c|}{\makecell{\colorone{64.83}\\\textcolor{gray}{\scriptsize$\pm$0.13}}}                           &  \multicolumn{2}{c}{\makecell{\colorone{72.27}\\\textcolor{gray}{\scriptsize$\pm$0.06}}}                           &  \\
  & Random Forest                      & \multicolumn{1}{l|}{} &  &     \multicolumn{2}{c|}{\makecell{\colorone{1.75}\\\textcolor{gray}{\scriptsize$\pm$0.00}}} & \multicolumn{2}{c|}{\makecell{\colorone{0.52}\\\textcolor{gray}{\scriptsize$\pm$0.00}}}  &  \multicolumn{2}{c|}{\makecell{\colorone{49.61}\\\textcolor{gray}{\scriptsize$\pm$0.06}}} &   \multicolumn{2}{c|}{\makecell{\colorone{67.82}\\\textcolor{gray}{\scriptsize$\pm$0.02}}}                           &  \multicolumn{2}{c}{\makecell{\colorone{74.46}\\\textcolor{gray}{\scriptsize$\pm$0.04}}}                           &  \\
   & ResNet-18                      & \multicolumn{1}{l|}{} &  &     \multicolumn{2}{c|}{\makecell{\colorone{1.79}\\\textcolor{gray}{\scriptsize$\pm$0.01}}} & \multicolumn{2}{c|}{\makecell{\colorone{0.51}\\\textcolor{gray}{\scriptsize$\pm$0.00}}}  &  \multicolumn{2}{c|}{\makecell{\colorone{48.85}\\\textcolor{gray}{\scriptsize$\pm$0.08}}} &   \multicolumn{2}{c|}{\makecell{\colorone{67.62}\\\textcolor{gray}{\scriptsize$\pm$0.09}}}                           &  \multicolumn{2}{c}{\makecell{\colorone{75.13}\\\textcolor{gray}{\scriptsize$\pm$0.16}}}                           &  \\\hline\hline
 & ResNet-18+FV                      & \multicolumn{1}{l|}{} &  &     \makecell{\colorone{1.79}\\\textcolor{gray}{\scriptsize$\pm$0.01}} & \multicolumn{1}{c|}{\makecell{\colortwo{1.21}\\\textcolor{gray}{\scriptsize$\pm$0.00}}}  &  \makecell{\colorone{0.51}\\\textcolor{gray}{\scriptsize$\pm$0.00}} &   \multicolumn{1}{c|}{\makecell{\colortwo{0.32}\\\textcolor{gray}{\scriptsize$\pm$0.00}}}                           &  \makecell{\colorone{48.85}\\\textcolor{gray}{\scriptsize$\pm$0.08}}                           &  
  \multicolumn{1}{c|}{\makecell{\colortwo{66.83}\\\textcolor{gray}{\scriptsize$\pm$0.08}}} & \makecell{\colorone{67.62}\\\textcolor{gray}{\scriptsize$\pm$0.09}}  &  \multicolumn{1}{c|}{\makecell{\colortwo{77.56}\\\textcolor{gray}{\scriptsize$\pm$0.05}}} &   \makecell{\colorone{75.13}\\\textcolor{gray}{\scriptsize$\pm$0.16}}                           &  \multicolumn{1}{c}{\makecell{\colortwo{78.13}\\\textcolor{gray}{\scriptsize$\pm$0.13}}}                           &\\
 & ResNet-18+HV                        & \multicolumn{1}{l|}{} &  &        \makecell{\colorone{1.79}\\\textcolor{gray}{\scriptsize$\pm$0.01}} & \multicolumn{1}{c|}{\makecell{\colortwo{1.67}\\\textcolor{gray}{\scriptsize$\pm$0.01}}}  &\makecell{\colorone{0.51}\\\textcolor{gray}{\scriptsize$\pm$0.00}}  &  \multicolumn{1}{c|}{\makecell{\colortwo{0.44}\\\textcolor{gray}{\scriptsize$\pm$0.00}}}  &\makecell{\colorone{48.85}\\\textcolor{gray}{\scriptsize$\pm$0.08}} &\multicolumn{1}{c|}{\makecell{\colortwo{55.20}\\\textcolor{gray}{\scriptsize$\pm$0.03}}}  &   \makecell{\colorone{67.62}\\\textcolor{gray}{\scriptsize$\pm$0.09}}                           & \multicolumn{1}{c|}{\makecell{\colortwo{71.31}\\\textcolor{gray}{\scriptsize$\pm$0.05}}}  & \makecell{\colorone{75.13}\\\textcolor{gray}{\scriptsize$\pm$0.16}}                           &  \multicolumn{1}{c}{\makecell{\colortwo{72.37}\\\textcolor{gray}{\scriptsize$\pm$0.10}}}                           &\\
 & ResNet-18+DE                        & \multicolumn{1}{l|}{} &  &     \makecell{\colorone{1.77}\\\textcolor{gray}{\scriptsize$\pm$0.01}} & \multicolumn{1}{c|}{\makecell{\colortwo{1.75}\\\textcolor{gray}{\scriptsize$\pm$0.06}}}  &  \makecell{\colorone{0.50}\\\textcolor{gray}{\scriptsize$\pm$0.00}} &   \multicolumn{1}{c|}{\makecell{\colortwo{0.48}\\\textcolor{gray}{\scriptsize$\pm$0.01}}}                           &  \makecell{\colorone{49.55}\\\textcolor{gray}{\scriptsize$\pm$0.01}}                           &  
  \multicolumn{1}{c|}{\makecell{\colortwo{51.77}\\\textcolor{gray}{\scriptsize$\pm$1.06}}} & \makecell{\colorone{68.11}\\\textcolor{gray}{\scriptsize$\pm$0.03}}  &  \multicolumn{1}{c|}{\makecell{\colortwo{69.40}\\\textcolor{gray}{\scriptsize$\pm$1.48}}} &   \makecell{\colorone{75.80}\\\textcolor{gray}{\scriptsize$\pm$0.01}}                           &  \multicolumn{1}{c}{\makecell{\colortwo{70.41}\\\textcolor{gray}{\scriptsize$\pm$0.56}}}                           &\\
 & ResNet-18+SE                        & \multicolumn{1}{l|}{} &  &     \makecell{\colorone{1.78}\\\textcolor{gray}{\scriptsize$\pm$0.00}} & \multicolumn{1}{c|}{\makecell{\colortwo{1.78}\\\textcolor{gray}{\scriptsize$\pm$0.00}}}  &  \makecell{\colorone{0.51}\\\textcolor{gray}{\scriptsize$\pm$0.00}} &   \multicolumn{1}{c|}{\makecell{\colortwo{0.51}\\\textcolor{gray}{\scriptsize$\pm$0.00}}}                           &  \makecell{\colorone{48.97}\\\textcolor{gray}{\scriptsize$\pm$0.03}}                           &  
  \multicolumn{1}{c|}{\makecell{\colortwo{49.39}\\\textcolor{gray}{\scriptsize$\pm$0.02}}} & \makecell{\colorone{67.68}\\\textcolor{gray}{\scriptsize$\pm$0.11}}  &  \multicolumn{1}{c|}{\makecell{\colortwo{67.88}\\\textcolor{gray}{\scriptsize$\pm$0.11}}} &   \makecell{\colorone{75.21}\\\textcolor{gray}{\scriptsize$\pm$0.16}}                           &  \multicolumn{1}{c}{\makecell{\colortwo{75.25}\\\textcolor{gray}{\scriptsize$\pm$0.17}}}                           &\\ 
  & ResNet-18+MCD                        & \multicolumn{1}{l|}{} &  &     \makecell{\colorone{1.83}\\\textcolor{gray}{\scriptsize$\pm$0.00}} & \multicolumn{1}{c|}{\makecell{\colortwo{1.78}\\\textcolor{gray}{\scriptsize$\pm$0.00}}}  &  \makecell{\colorone{0.52}\\\textcolor{gray}{\scriptsize$\pm$0.00}} &   \multicolumn{1}{c|}{\makecell{\colortwo{0.49}\\\textcolor{gray}{\scriptsize$\pm$0.00}}}                           &  \makecell{\colorone{47.56}\\\textcolor{gray}{\scriptsize$\pm$0.10}}                           &  
  \multicolumn{1}{c|}{\makecell{\colortwo{49.59}\\\textcolor{gray}{\scriptsize$\pm$0.15}}} & \makecell{\colorone{66.73}\\\textcolor{gray}{\scriptsize$\pm$0.01}}  &  \multicolumn{1}{c|}{\makecell{\colortwo{67.85}\\\textcolor{gray}{\scriptsize$\pm$0.03}}} &   \makecell{\colorone{74.69}\\\textcolor{gray}{\scriptsize$\pm$0.04}}                           &  \multicolumn{1}{c}{\makecell{\colortwo{75.04}\\\textcolor{gray}{\scriptsize$\pm$0.03}}}                           &\\ 
  & ResNet-18+MVN                        & \multicolumn{1}{l|}{} &  &     \makecell{\colorone{1.94}\\\textcolor{gray}{\scriptsize$\pm$0.03}} & \multicolumn{1}{c|}{\makecell{\colortwo{1.14}\\\textcolor{gray}{\scriptsize$\pm$0.01}}}  &  \makecell{\colorone{0.55}\\\textcolor{gray}{\scriptsize$\pm$0.01}} &   \multicolumn{1}{c|}{\makecell{\colortwo{0.25}\\\textcolor{gray}{\scriptsize$\pm$0.00}}}                           &  \makecell{\colorone{45.16}\\\textcolor{gray}{\scriptsize$\pm$0.22}}                           &  
  \multicolumn{1}{c|}{\makecell{\colortwo{68.77}\\\textcolor{gray}{\scriptsize$\pm$0.11}}} & \makecell{\colorone{64.89}\\\textcolor{gray}{\scriptsize$\pm$0.29}}  &  \multicolumn{1}{c|}{\makecell{\colortwo{77.94}\\\textcolor{gray}{\scriptsize$\pm$0.11}}} &   \makecell{\colorone{72.61}\\\textcolor{gray}{\scriptsize$\pm$0.29}}                           &  \multicolumn{1}{c}{\makecell{\colortwo{76.50}\\\textcolor{gray}{\scriptsize$\pm$0.28}}}                           &\\ 
  & ResNet-18+HetReg                    & \multicolumn{1}{l|}{} &  &     \makecell{\colorone{1.94}\\\textcolor{gray}{\scriptsize$\pm$0.00}} & \multicolumn{1}{c|}{\makecell{\colortwo{1.14}\\\textcolor{gray}{\scriptsize$\pm$0.00}}}  &  \makecell{\colorone{0.55}\\\textcolor{gray}{\scriptsize$\pm$0.00}} &   \multicolumn{1}{c|}{\makecell{\colortwo{0.26}\\\textcolor{gray}{\scriptsize$\pm$0.00}}}                           &  \makecell{\colorone{44.97}\\\textcolor{gray}{\scriptsize$\pm$0.02}}                           &  
  \multicolumn{1}{c|}{\makecell{\colortwo{68.60}\\\textcolor{gray}{\scriptsize$\pm$0.04}}} & \makecell{\colorone{64.88}\\\textcolor{gray}{\scriptsize$\pm$0.11}}  &  \multicolumn{1}{c|}{\makecell{\colortwo{77.86}\\\textcolor{gray}{\scriptsize$\pm$0.01}}} &   \makecell{\colorone{72.49}\\\textcolor{gray}{\scriptsize$\pm$0.03}}                           &  \multicolumn{1}{c}{\makecell{\colortwo{76.33}\\\textcolor{gray}{\scriptsize$\pm$0.02}}}                           &\\ 
  \hline 

\end{tabular}
}
\end{subtable}
\end{minipage}
\caption{Performance of SDM-estimation approaches both \colorone{without Bayesian updates} (left side in split columns) and \colortwo{after updates from five checklists} (right side). Note that the first four methods for each region are not uncertainty-aware and therefore cannot be updated with checklist information. Means are shown $\pm$ standard deviation across runs. (Note that many standard deviations are quite low; those reported as 0.00 are accurate.)}
    \label{resusawinter}
\end{table*}

\begin{figure*}[h!]
    \centering
\includegraphics[width=0.99\linewidth]{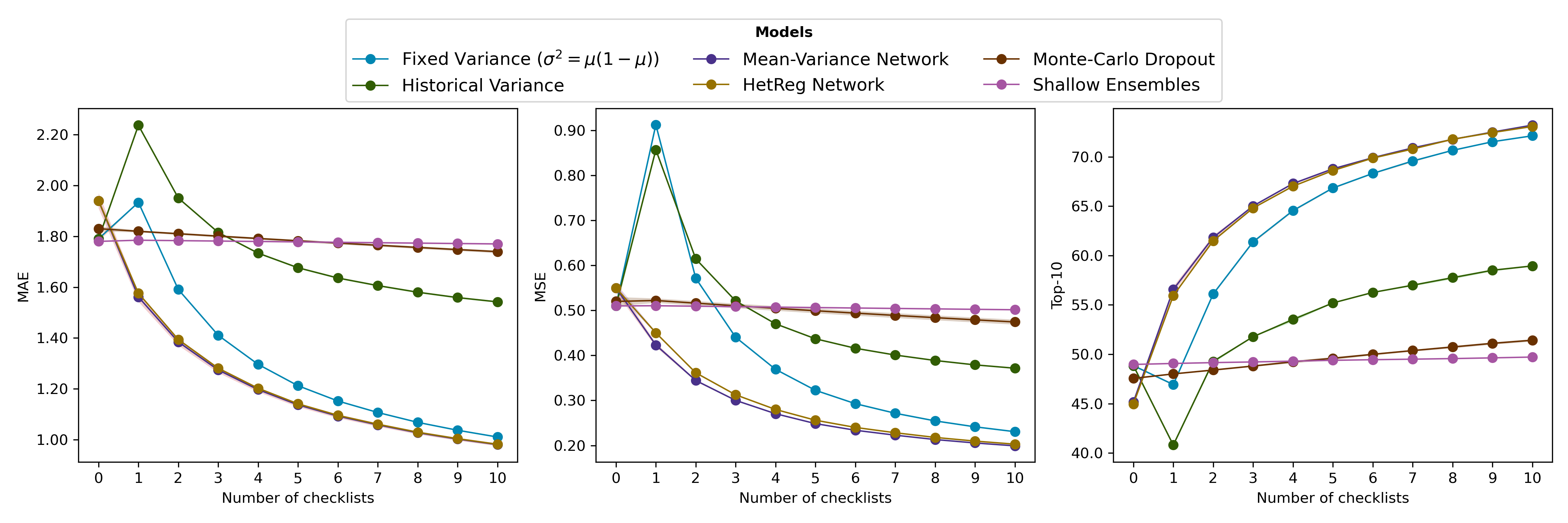}
\caption{Iterative improvements for the different uncertainty estimation approaches with increasing
number of checklist updates for the MAE, MSE and Top-10 metrics on the USA-Winter Region test
set. We report the mean on three seeds and standard deviations for each model.}
\label{fig:usawintergraphsappendix}
\end{figure*}

\clearpage



\end{document}